\documentclass[preprint,12pt,authoryear]{elsarticle}
\journal{Insurance: Mathematics and Economics}

\usepackage{graphicx}
\usepackage{subcaption}
\usepackage{amsmath,amssymb}
\usepackage{tikz}
\usetikzlibrary{arrows.meta,positioning}
\usetikzlibrary{decorations.pathreplacing}
\usepackage{tcolorbox}
\usepackage{booktabs}
\usepackage{multirow}
\usepackage{hyperref}
\usepackage{natbib}
\usepackage[margin=1.0in]{geometry}
\usepackage{xcolor}
\usepackage{rotating}
\usepackage{tabularx}

\definecolor{tgnblue}{RGB}{38, 93, 171}
\definecolor{tgnbluelight}{RGB}{222, 235, 247}
\definecolor{darkgreen}{RGB}{0,150,0}

\newcommand{\lantern}{\textsc{LANTERN}}

\begin{document}
\begin{frontmatter}



\title{A Horizon-Specific Attribute-Conditioned Neural Network for Estimating Health-State Transition Probabilities in Irregular Longitudinal Data: The \lantern{} Framework}

\author[1]{Bright Kwaku Manu\corref{cor1}}
\ead{bkmanu@asu.edu}
\author[2]{Beckett Sterner}
\author[3]{Petar Jevti\'c}

\address[1]{School of Computing and Augmented Intelligence, Arizona State University, Tempe, USA}
\address[2]{School of Life Sciences, Arizona State University, Tempe, USA}
\address[3]{School of Mathematical and Statistical Sciences, Arizona State University, Tempe, USA}

\cortext[cor1]{Corresponding author}

\begin{abstract}
Accurate estimation of health-state transition probabilities is essential for long-term care insurance modeling. Conventional actuarial multi-state models provide an interpretable framework but may be restrictive in capturing nonlinear relationships and accumulated health histories from irregular longitudinal observations. This paper proposes LANTERN, a history-dependent neural framework that integrates recurrent health representations, elapsed observation time, and adaptive demographic and socioeconomic attribute conditioning to estimate horizon-specific transition probabilities among Healthy, Mild disability, Severe disability, and Death states. 
Using longitudinal Health and Retirement Study data, we evaluate our model against a recurrent neural network, multinomial logistic regression, gradient-boosted trees, and an actuarial continuous-time Markov chain across one-, two-, and three-year forecast horizons. At the primary two-year horizon, LANTERN achieves competitive predictive accuracy and calibration, with improvements over selected benchmarks. Its aggregated transition-matrix estimates outperform multinomial logistic regression and are comparable to those of the strongest remaining benchmarks. These findings suggest that history-conditioned neural estimation provides a flexible and competitive approach to estimating horizon-specific health-state transition probabilities, with potential applications in actuarial long-term care modeling.
\end{abstract}


\end{frontmatter}

\section{Introduction}
\label{sec:introduction}

Long-term care (LTC) risk is becoming an increasingly important source of financial uncertainty for households, insurers, and public programs. As populations age, more individuals are expected to spend part of later life with functional limitations that require formal or informal care. LTC insurance products are intended to pool and pre-fund part of this risk, but their viability depends on credible estimates of how individuals move among functional health states over time \cite{pitacco2014health,sherris2021multi,park2024design}.

In actuarial LTC models, disability progression and mortality are commonly represented using finite-state models. Individuals move between states such as healthy, mildly disabled, severely disabled, and dead, and the associated transition probabilities determine projected disability prevalence, expected benefit payments, reserves, and solvency requirements \cite{haberman1997multiple,fong2015multistate,pitacco2014health,sherris2021multi}. Under classical multi-state cohort projection, repeated application of transition matrices over future ages can propagate estimation errors, potentially affecting long-term valuation results \cite{christiansen2010biometric,maegebier2013valuation}. For actuarial applications, these transition probabilities must correspond to a clearly specified future interval or future age. We refer to this future interval as the \emph{prediction horizon}.

Estimating these transition probabilities is difficult in longitudinal aging data as future functional status may depend not only on the current disability state, but also on accumulated health history, previous disability episodes, chronic conditions, demographic and socioeconomic characteristics, and the timing of earlier observations \cite{fong2019disability,soutinho2020estimation}. Longitudinal health data therefore contain information that extends beyond the most recent health state alone. In surveys such as the Health and Retirement Study, repeated observations are collected primarily on a biennial schedule, but realized observation intervals vary because respondents may miss waves and because interview timing differs across individuals \cite{bugliari2025rand}. The resulting modeling problem is therefore not simply one of variable observation timing, but of learning from longitudinal histories observed at different intervals while estimating transition probabilities over a clearly specified prediction horizon.


Classical actuarial and survival models provide interpretable transition structures and a direct connection to valuation, but they can be restrictive when transition risk depends on nonlinear relationships, heterogeneous covariate effects, and accumulated longitudinal health history. Recent work has begun to connect machine learning with health transition and multi-state survival modeling. For example, \cite{wang2022multistate} combine neural networks with a generalized linear model to estimate and predict health transition intensities, allowing socioeconomic and lifestyle factors to enter through linear and nonlinear relationships. In multi-state survival analysis, \cite{rahman2023multi} propose pseudo-value-based deep neural networks for subject-specific prediction of multi-state quantities, including transition probabilities and state occupation probabilities, in the presence of censoring. More broadly, machine learning methods have been used for longitudinal clinical prediction tasks such as mortality, readmission, length-of-stay prediction, physiologic decompensation, and medical state pre-warning \cite{rajkomar2018scalable,harutyunyan2019multitask,nie2022forecasting}. For LTC insurance product applications, however, a useful model must do more than rank individuals by risk. It must produce calibrated probability vectors over the possible health states, so that individual-level predictions can be aggregated into transition matrices for cohort projection \cite{guo2017calibration,steyerberg2014towards,denuit2021autocalibration}. An important methodological question is whether flexible longitudinal representations can be combined with explicitly specified actuarial prediction horizons while retaining calibrated multi-state probability outputs.


As a contribution to this important research topic, this paper develops \lantern{} (\textbf{L}ongitudinal \textbf{A}ttribute-conditioned \textbf{N}eural \textbf{T}ransition \textbf{E}stimation \textbf{R}ecurrent \textbf{N}etwork), a history-dependent framework for estimating multi-state health transition probabilities at specified prediction horizons. The central idea of this work is to retain the actuarial multi-state probability framework while allowing transition risk to depend on a learned representation of accumulated longitudinal health history and an explicitly specified future horizon. The contribution of the paper is fourfold. First, we formulate LTC transition estimation as a history-dependent, horizon-specific forecasting problem rather than a next-observation prediction problem. Second, we develop a recurrent health-history representation that summarizes repeated functional, clinical, and health-related information while accounting for the temporal spacing of previous observations. Third, we introduce a history-conditioned attribute-attention mechanism that allows the relative weighting of demographic and socioeconomic representations to vary with an individual's evolving health history, while providing a diagnostic of the model's internal attribute allocation. Fourth, we integrate the prospective forecast horizon into the prediction head, separately from historical observation timing, to obtain coherent multi-state transition probability vectors that can be aggregated into age- and origin-state-specific transition matrices.


Using longitudinal data from the Health and Retirement Study, we compare \lantern{} with an age- and sex-dependent continuous-time actuarial multi-state model, multinomial logistic regression, gradient-boosted trees, and a recurrent neural-network benchmark. Because the observed HRS follow-up distribution is concentrated near the nominal biennial interval, the two-year horizon is used as the primary actuarial evaluation setting, while additional horizons are examined where sufficient empirical support is available. The empirical analysis evaluates both individual-level probabilistic performance and fixed-horizon aggregate transition accuracy. This distinction is important because an estimator useful for actuarial applications must not only distinguish higher-risk from lower-risk individuals, but also produce horizon-specific transition probabilities whose accuracy and calibration can be evaluated at both individual and aggregate levels.

The remainder of the paper is organized as follows. Section~\ref{relwork} reviews related work on actuarial multi-state modeling, transition probability estimation, machine learning, and calibration. Section~\ref{class_actform} presents the classical actuarial multi-state framework. Section~\ref{sec:data} describes the data and empirical setting. Section~\ref{method} introduces the proposed \lantern{} framework and its estimation methodology. Section~\ref{exp} describes the training procedure, evaluation metrics, and benchmark models. Section~\ref{sec:results} presents the empirical findings, while Section~\ref{sec:discussion} discusses their implications and limitations. Finally, Section~\ref{sec:conclusion} concludes.

\section{Related Work}
\label{relwork}

\subsection{Actuarial Multi-State Models for Long-Term Care}

Actuarial modeling of long-term care (LTC) and disability risk commonly relies on multi-state models representing transitions among functional states such as healthy, mild disability, severe disability, and death, with death treated as absorbing. These models accommodate disability progression, recovery, and mortality, providing a basis for disability transition tables, dependence probability tables, LTC insurance valuation, and care-annuity modeling \cite{haberman1997multiple,fong2015multistate,pitacco2014health,sherris2021multi}. Transition dynamics may be represented through continuous-time intensities or discrete-time, age-specific transition probability matrices. In actuarial cohort projection, these matrices are applied over successive intervals to estimate future state occupancy and associated benefit payments, reserves, and solvency requirements. Consequently, transition-probability errors may accumulate over long projection horizons and affect insurance valuation \cite{haberman1997multiple,christiansen2010biometric,maegebier2013valuation}.

Multi-state LTC models have been used to estimate expected care time, healthy life expectancy, disability prevalence, and the financial consequences of disability progression. Extensions incorporate trends, parameter uncertainty, information delays, and longevity-related risks \cite{levantesi2012managing,sherris2021multi,sandqvist2023multistate}. When longitudinal transition histories are unavailable, transition probabilities can also be inferred from repeated cross-sectional prevalence and mortality data under Markov assumptions \cite{fong2015multistate,kessy2024estimating}.

The limitation to overcome considered in this study is not the actuarial multi-state framework itself, but the specification used to estimate its transition probabilities. Conventional Markov formulations commonly condition transition risk on current state and age, while semi-Markov and covariate-extended models can incorporate duration and observed risk factors. Although these extensions offer substantial flexibility, they generally require the analyst to specify which aspects of prior health history enter the transition mechanism. This motivates investigating whether a learned longitudinal representation can support fixed-horizon transition-probability estimation while preserving the multi-state structure needed for actuarial analysis.

\subsection{Estimating Transition Probabilities}

Transition-probability estimation also arises in health economic Markov cohort models, where transition matrices describe movement among disease or care states and formal guidance on their estimation remains limited \cite{olariu2017current}. Existing approaches differ in their assumptions about transition dynamics, covariate effects, longitudinal history, and functional-form flexibility rather than forming mutually exclusive categories. We review them with emphasis on how they incorporate prior health information, define prediction horizons, and estimate multi-state transition probabilities.

\paragraph{\it Classical and Survival-Based Estimation}
Classical approaches estimate transition probabilities using parametric, non-parametric, and semi-parametric multi-state methods. In survival analysis, transition intensities and cumulative hazards are studied using counting-process methods, likelihood-based estimation, and Nelson--Aalen-type estimators, while transition probabilities can be estimated using the Aalen--Johansen estimator under appropriate assumptions \cite{andersen1993statistical,hougaard1999multi,klein2003survival,kalbfleisch2011statistical}. These methods provide a rigorous inferential foundation and have been extended to competing risks, illness-death models, time-dependent covariates, semi-Markov settings, and non-Markov transition-probability estimation \cite{d2009full,guibert2018non,soutinho2020estimation,llopis2023estimating}.

For Markov processes, the Aalen--Johansen estimator is widely used to estimate transition and state occupation probabilities. However, conditioning only on current state may not adequately capture accumulated frailty, previous disability episodes, or duration-dependent risk when the Markov assumption is violated \cite{d2009full,guibert2018non,coemans2022bias}. Non-Markov, semi-Markov, duration-dependent, and landmarking approaches address aspects of this limitation by incorporating previous states, sojourn time, elapsed time, and intermediate events \cite{soutinho2020estimation}. These developments demonstrate that history dependence and prediction over specified future intervals are well established in survival and multi-state modeling \cite{soutinho2020estimation,d2009full,guibert2018non,llopis2023estimating}. Their practical implementation, however, generally requires explicit choices about the relevant history summaries, duration measures, and transition structure.

\paragraph{\it GLM-Based Estimation}
Generalized linear models (GLMs) provide interpretable specifications for estimating health-state transitions. In actuarial applications, they accommodate covariates such as age, sex, duration, time trends, demographic characteristics, clinical risk factors, and interaction effects \cite{fong2015multistate,hanewald2019modelling,levantesi2021modelling,tanasia2024generalized,curioso2025risk}. Transition probabilities can be estimated directly in discrete time or derived from continuous-time transition intensities through Kolmogorov forward equations \cite{christiansen2010biometric,olariu2017current}.

Discrete-time specifications include logistic, multinomial logistic, ordinal, proportional-odds, and complementary log-log models, depending on the transition structure and outcome \cite{olariu2017current,kalbfleisch2011statistical,klein2003survival}. For example, multinomial logistic regression estimates origin-state-specific probabilities across possible destinations while ensuring that each probability vector sums to one, allowing the estimates to be assembled into transition matrices. GLM-based models provide an important benchmark because they can represent nonlinear effects, elapsed time, and covariate interactions when appropriately specified. Frailty and random-effect formulations further accommodate latent individual- or group-level heterogeneity \cite{wienke2010frailty}. The principal distinction from the present approach is therefore not an inability to represent time or heterogeneity, but how longitudinal history is summarized. In these models, the analyst typically specifies the relevant previous states, duration measures, covariates, latent effects, and interaction terms in advance.

\paragraph{\it Machine Learning-Based Health-State Modeling}
Machine learning offers flexible alternatives for representing nonlinear relationships and longitudinal health information. A particularly relevant actuarial contribution is \cite{wang2022multistate}, who combine neural networks with a generalized linear modeling framework to estimate age-specific transition intensities in a multi-state Markov model. Their approach incorporates nonlinear socioeconomic and lifestyle effects and uses transfer learning across transition processes. In contrast, LANTERN uses recurrent memory to summarize accumulated longitudinal health history and directly estimates multi-state transition probabilities at specified future horizons, distinguishing historical observation timing from the prospective prediction interval.

Neural methods have also been developed for multi-state survival analysis. For example, \cite{rahman2023multi} use pseudo-value-based deep neural networks to predict subject-specific quantities, including transition and state occupation probabilities, in the presence of censoring. Continuous-time hidden Markov models address irregular clinical observations through latent disease states and continuous-time transitions \cite{liu2015efficient}. These approaches provide flexible tools for multi-state and disease-progression modeling, although their objectives differ from constructing observed-state transition matrices for actuarial LTC applications.

In broader healthcare applications, deep learning has been applied to longitudinal electronic health records and clinical time series. \cite{rajkomar2018scalable} predict outcomes including mortality, readmission, length of stay, and discharge diagnoses, while \cite{harutyunyan2019multitask} develop benchmarks for mortality, physiological decompensation, length-of-stay prediction, and phenotype classification. Short-term medical state forecasting has also been investigated using logistic regression, AdaBoost, and XGBoost \cite{nie2022forecasting}. Although these studies demonstrate the predictive value of complex health histories, they do not directly address calibrated, fixed-horizon multi-state transition estimation for actuarial LTC analysis. For actuarial applications, predictive flexibility must therefore be accompanied by reliable probability estimates that can be evaluated and aggregated within a multi-state framework.

\subsection{Calibration of Probabilistic Predictions}

Calibration describes the agreement between predicted probabilities and observed event frequencies. It is particularly relevant to LTC transition modeling as estimated probabilities may be aggregated into transition matrices for actuarial analysis. Systematic overestimation or underestimation can distort projected state occupancy and subsequent valuation outputs \cite{steyerberg2014towards,van2019calibration,eddy2012model}. High predictive accuracy does not necessarily imply well-calibrated probabilities. \cite{guo2017calibration} demonstrate that neural networks can achieve strong classification performance while producing miscalibrated predictions and examine post-processing methods such as temperature scaling. Calibration assessment therefore complements discrimination measures and may include reliability diagrams, calibration-in-the-large, calibration slope, Brier score, and Expected Calibration Error \cite{brier1950verification,steyerberg2014towards,guo2017calibration}. In multi-state settings, calibration must account for probabilities across multiple destination states rather than a single binary endpoint.

Actuarial research has also emphasized calibration and auto-calibration as desirable properties of predictive models used in insurance applications \cite{denuit2021autocalibration}. Accordingly, LTC transition estimators should be evaluated in terms of probabilistic accuracy, calibration, and the fidelity of their aggregated fixed-horizon transition probabilities, rather than discrimination alone.

Taken together, the literature provides established methods for modeling multi-state transitions, incorporating covariates and duration effects, and representing longitudinal history. The question addressed here is whether learned recurrent representations can combine accumulated health history, irregular observation timing, and demographic and socioeconomic information to estimate calibrated multi-state transition probabilities at explicitly specified future horizons.

\section{Classical Actuarial Multi-State Framework}
\label{class_actform}

The estimated transition probabilities in this study are intended for use in standard actuarial multi-state projection frameworks \cite{haberman1997multiple,pitacco2014health}. We briefly review this framework and define the transition probability notation used throughout the analysis.

Consider a finite state space $\mathcal{J}=\{H,M,S,D\}$, where $H$, $M$, and $S$ denote healthy, mild disability, and severe disability states, respectively, and $D$ denotes death as an absorbing state. Let $Y(a)$ represent an individual's health state at age $a$. For a specified prediction horizon $\tau>0$, define the transition probability
\begin{equation}
\label{eq:classical-transition-prob}
p_{rs}(a, \tau)=\Pr\{Y(a+\tau)=s\mid Y(a)=r\}, \qquad r,s\in\mathcal{J}.
\end{equation}
where $r$ and $s$ denote the origin and destination states in $\mathcal{J}$, respectively. Figure~\ref{fig:ltc_markov_4state} illustrates the four-state LTC transition structure at age $a$

\begin{figure}[h]
\centering
\begin{tikzpicture}[
    state/.style={
        circle,
        draw=black,
        thick,
        minimum size=1.05cm,
        align=center
    },
    trans/.style={
        -{Latex[length=2mm]},
        thick
    },
    prob/.style={
        font=\scriptsize,
        fill=white,
        inner sep=1pt
    }
]

\node[state] (H) at (0,2.5) {$H$};
\node[state] (M) at (-2.8,0) {$M$};
\node[state] (S) at (2.8,0) {$S$};
\node[state] (D) at (0,-2.5) {$D$};

\draw[trans] (H) to[bend right=10]
    node[prob, pos=0.45, midway, above left] {$p_{HM}(a,\tau)$} (M);
\draw[trans] (M) to[bend right=10]
    node[prob, pos=0.35, above right] {$p_{MH}(a,\tau)$} (H);

\draw[trans] (H) to[bend left=10] (S);
\draw[trans] (S) to[bend left=10] (H);

\draw[trans] (M) to[bend left=8]
    node[prob, pos=0.55, above right] {$p_{MS}(a,\tau)$} (S);
\draw[trans] (S) to[bend left=8] (M);

\draw[trans] (H) to[bend right=6] (D);
\draw[trans] (M) to[bend right=10] (D);
\draw[trans] (S) to[bend left=10]
    node[prob, pos=0.55, right] {$p_{SD}(a,\tau)$} (D);

\draw[trans] (D) to[loop below, min distance=8mm]
    node[prob] {$p_{DD}(a,\tau)=1$} (D);

\draw[trans] (H) to[loop above, min distance=8mm]
    node[prob] {$p_{HH}(a,\tau)$} (H);

\draw[trans] (M) to[loop left, min distance=8mm]
    node[prob] {$p_{MM}(a,\tau)$} (M);

\draw[trans] (S) to[loop right, min distance=8mm]
    node[prob] {$p_{SS}(a,\tau)$} (S);

\end{tikzpicture}
\caption{\textbf{Four-state LTC transition structure.} States \(H\), \(M\), \(S\), and \(D\) denote healthy, mild disability, severe disability, and death, respectively. Arrows represent horizon-specific transition probabilities \(p_{rs}(a,\tau)\). Persistence within each living state is permitted through \(p_{HH}(a,\tau)\), \(p_{MM}(a,\tau)\), and \(p_{SS}(a,\tau)\). Death is absorbing, with \(p_{DD}(a,\tau)=1\).}
\label{fig:ltc_markov_4state}
\end{figure}

The corresponding transition matrix is denoted by $\mathbf{P}(a,\tau)=\big(p_{rs}(a,\tau)\big)_{r,s\in\mathcal{J}}$. Each row of $\mathbf{P}(a,\tau)$ is a probability vector, so $p_{rs}(a,\tau)\geq 0$ and $\sum_{s\in\mathcal{J}}p_{rs}(a,\tau)=1$. Persistence within the living states is permitted through $p_{HH}(a,\tau)$, $p_{MM}(a,\tau)$, and $p_{SS}(a,\tau)$. Since death is absorbing, $p_{DD}(a,\tau)=1$ and $p_{Ds}(a,\tau)=0$ for $s\neq D$.

Let $\boldsymbol{\pi}_a$ denote the row vector of state occupancy probabilities at age $a$. Under the classical multi-state projection framework, cohort occupancy over a specified prediction horizon $\tau$ is given by as \cite{haberman1997multiple,pitacco2014health}
\begin{equation}
\label{eq:cohort-recursion}
\boldsymbol{\pi}_{a+\tau}
=
\boldsymbol{\pi}_a\mathbf{P}(a,\tau).
\end{equation}

Under the Markov assumption, repeated application of age-specific transition matrices yields the occupancy vector after $k$ projection intervals as
\begin{equation}
\boldsymbol{\pi}_{a_0+k\tau}
=
\boldsymbol{\pi}_{a_0}
\mathbf{P}(a_0,\tau)
\mathbf{P}(a_0+\tau,\tau)
\cdots
\mathbf{P}(a_0+(k-1)\tau,\tau).
\end{equation}

Errors in estimated transition probabilities may propagate through successive projection steps due to repeated matrix multiplication, potentially affecting projected disability prevalence, expected benefit payments, reserves, and other actuarial quantities \cite{haberman1997multiple,christiansen2010biometric,maegebier2013valuation}.

Classical implementations commonly assume first-order Markov dependence on observed states, semi-Markov dependence on duration in state, or parametric covariate effects within proportional-hazard or transition-intensity structures \cite{haberman1997multiple,levantesi2021modelling,soutinho2020estimation}. These approaches provide the actuarial multi-state structure required for projection, but the transition mechanism is typically specified through a predefined dependence structure. The proposed methodology retains the row-stochastic multi-state probability structure while allowing horizon-specific transition risk to depend on accumulated longitudinal health history, demographic and socioeconomic attributes, and the timing of previously observed health information. Thus, the proposed method provides an alternative approach to estimating age- and horizon-specific transition matrices within the classical actuarial multi-state structure.

\section{Data and Empirical Setting}
\label{sec:data}

\subsection{Data Source and Longitudinal Structure}

We use the RAND Health and Retirement Study (HRS) Longitudinal File 2022 (V1), restricting the empirical analysis to survey waves from 1992 through 2022 \cite{bugliari2025rand}. The HRS is a nationally representative longitudinal study of U.S. adults conducted primarily on a biennial schedule. Because respondents may miss scheduled interviews or be observed at different ages across waves, realized follow-up intervals vary. We therefore treat each retained person-wave record as a visit and use age at interview as the longitudinal time index.

The original RAND respondent-level file contains 45,234 individuals. Wave-specific variables are reshaped into a person-wave structure, observations with missing interview age are excluded, and the analytic age range is restricted to 30--100 years. Living observations with missing ADL or IADL information are also excluded from health-state classification. After retaining records with a valid subsequent state and positive forecast horizon, the analytic sample contains 227,370 transition records from 37,713 individuals. Among these individuals, the retained longitudinal histories comprise 2--15 visits per respondent, with a mean of 7.03 (SD = 3.65) and a median of 7 visits (interquartile range: 4--10). The distribution of visits is reported in Supplementary Table \ref{tab:visit_distribution}. Visit counts are calculated before the final transition eligibility filter, whereas transition counts include only valid origin--destination pairs. Although the HRS primarily samples adults aged 50 and older, younger observations may arise because spouses or partners may also be interviewed. Age-specific transition counts and accumulated inter-visit follow-up years are reported in Supplementary Table \ref{tab:age_exposure} to characterize the empirical support across age groups.

\subsection{Health-State Construction}
Functional health status is constructed from the six-item Activities of Daily Living (ADL) index and five-item Instrumental Activities of Daily Living (IADL) index. We consider the four-state space $\mathcal{J}=\{H,M,S,D\}$, representing Healthy, Mild disability, Severe disability, and Death. For an individual observed alive at visit $v$,
$$
\begin{aligned}
H &: \mathrm{ADL}_{iv}=0,\quad \mathrm{IADL}_{iv}=0,\\
M &: \mathrm{ADL}_{iv}=1
     \quad \text{or}\quad
     \left(\mathrm{ADL}_{iv}=0,\ \mathrm{IADL}_{iv}\geq1\right),\\
S &: \mathrm{ADL}_{iv}\geq2.
\end{aligned}
$$
Living observations with missing ADL or IADL information are excluded from state classification rather than assigned to a health state through imputation. Death is treated as an absorbing state using recorded age at death. A retained record is classified as dead when age at death is available and the corresponding observed age is at or beyond the recorded death age. The resulting health state defines the origin and destination states used in transition analysis. It is not supplied separately to the neural model because the ADL and IADL measures from which it is constructed are already included in the longitudinal health information.

\subsection{Covariates and Preprocessing}

The longitudinal feature set includes functional status, self-rated health, cognition, depressive symptoms, body mass index, chronic conditions, marital status, and census region. Chronic conditions include diabetes, cancer, lung disease, heart disease, stroke, and psychiatric conditions. These are represented as binary indicators and summed to form a disease-burden measure.

Demographic and socioeconomic attributes include sex, race, Hispanic ethnicity, education, marital status, and census region. Sex, race, Hispanic ethnicity, and the RAND cross-wave education measures are treated as respondent-level characteristics, whereas marital status and census region may vary across visits and are evaluated at the corresponding observation.

Missingness indicators are constructed for longitudinal variables with incomplete responses. Changes between consecutive retained visits are also calculated for selected health measures, including ADL, IADL, cognition, depressive symptoms, body mass index, and disease burden. 
Detailed variable definitions and coding are provided in Supplementary Table~\ref{tab:rand_variables}.

\subsection{Observation Timing and Forecast Horizons}
For each retained origin visit $v$, the next retained observation defines the destination state and realized forecast horizon,
$$
\tau_i^{(v)} = a_i^{(v+1)} - a_i^{(v)}.
$$
The corresponding observed destination is $Y_i^{(v+1)}$. Records without a subsequent retained observation or with $\tau_i^{(v)}\leq0$ are excluded.

As shown in Table~\ref{tab:horizon_distribution}, realized follow-up is concentrated around the nominal biennial HRS schedule, with 71.5\% of retained transitions occurring over two years. The mean and median forecast horizons are 2.09 and 2 years, respectively. Horizons range from 1 to 21 years, although observations beyond three years are comparatively sparse.
\begin{table}[h]
\centering
\caption{Distribution of realized forecast horizons in the analytic transition sample.}
\label{tab:horizon_distribution}
\begin{tabular}{lrr}
\hline
\textbf{Forecast horizon} & \textbf{Transitions} & \textbf{Percent} \\
\hline
1 year   & 29,055  & 12.8 \\
2 years  & 162,491 & 71.5 \\
3 years  & 29,429  & 12.9 \\
4 years  & 3,277   & 1.4 \\
5+ years & 3,118   & 1.4 \\
\hline
Total    & 227,370 & 100.0 \\
\hline
\end{tabular}
\end{table}
This distribution motivates the empirical evaluation strategy. The two-year horizon is treated as the primary actuarial setting because it has the strongest empirical support and corresponds most closely to the intended HRS interview schedule. One- and three-year horizons are considered as additional supported settings. Longer horizons may contribute to horizon-aware model fitting but are not treated as equally supported actuarial evaluation points because of their limited sample sizes.

\section{Methods}
\label{method}

\subsection{Problem Formulation}
\label{subsec:problem}

The classical projection framework in Section~\ref{class_actform} is expressed in terms of age- and horizon-specific transition probabilities $p_{rs}(a,\tau)$. In the longitudinal data used here, individuals are observed at a sequence of survey visits occurring at potentially unequal ages rather than on an exactly regular time grid.

Let $N$ denote the number of individuals indexed by $i\in\{1,\dots,N\}$. Individual $i$ is observed at visits $v=1,\dots,V_i$, ordered by survey wave and occurring at ages
$$
a_i^{(1)}<a_i^{(2)}<\cdots<a_i^{(V_i)},
$$
where $V_i$ is the number of observed visits. Visits are processed in survey-wave order, while age is used to measure elapsed time between observations.

At visit $v$, we observe a time-varying health-related feature vector $\mathbf{x}_i^{(v)}\in\mathbb{R}^{d_x}$ and a vector of demographic and socioeconomic attributes $\mathbf{c}_i^{(v)}=(c_{i1}^{(v)},\dots,c_{iA}^{(v)})$. The health-related features include functional status, clinical conditions, comorbidity measures, and other longitudinal health covariates. The attribute vector includes sex, race, Hispanic ethnicity, education, marital status, and region; some attributes are respondent-level, whereas others may vary across visits.

Let $Y_i^{(v)}\in\mathcal{J}$ denote the observed health state at visit $v$, where $\mathcal{J}=\{H,M,S,D\}$. The state variable defines the origin and destination states used in transition analysis and subsequent actuarial aggregation. It is not supplied as a separate model input because it is constructed from the ADL and IADL measures already contained in $\mathbf{x}_i^{(v)}$.

To represent historical observation timing, for $v\geq2$ define
$$
\Delta a_i^{(v)}=a_i^{(v)}-a_i^{(v-1)},
$$
with $\Delta a_i^{(1)}=0$. The quantity $\Delta a_i^{(v)}$ describes the elapsed time between consecutive observed visits and is known when visit $v$ is observed. Figure~\ref{fig:timing_structure} illustrates the observed visit sequence and distinguishes these historical elapsed intervals from the prospective prediction horizon introduced below.

\begin{figure*}[t]
\centering
\begin{tikzpicture}[
    >=Latex,
    font=\small,
    visit/.style={
        draw,
        minimum width=1.20cm,
        minimum height=1.35cm,
        align=center
    }
]

\node[visit] (v1) at (0,0)
{$\mathbf{c}_i^{(1)}$\\[3pt]$\mathbf{x}_i^{(1)}$};

\node[visit] (v2) at (2.15,0)
{$\mathbf{c}_i^{(2)}$\\[3pt]$\mathbf{x}_i^{(2)}$};

\node[visit] (v3) at (5.15,0)
{$\mathbf{c}_i^{(3)}$\\[3pt]$\mathbf{x}_i^{(3)}$};

\node at (6.75,0) {$\cdots$};

\node[visit] (vm1) at (8.55,0)
{$\mathbf{c}_i^{(v-1)}$\\[3pt]$\mathbf{x}_i^{(v-1)}$};

\node[visit] (v) at (10.85,0)
{$\mathbf{c}_i^{(v)}$\\[3pt]$\mathbf{x}_i^{(v)}$};

\draw[->] (v1.east) -- node[above, midway] {$\Delta a_i^{(2)}$} (v2.west);
\draw[->] (v2.east) -- node[above, midway] {$\Delta a_i^{(3)}$} (v3.west);
\draw[->] (vm1.east) -- node[above, midway] {$\Delta a_i^{(v)}$} (v.west);

\node[above=0.28cm of v1] {$v=1$};
\node[above=0.28cm of v2] {$v=2$};
\node[above=0.28cm of v3] {$v=3$};
\node[above=0.28cm of vm1] {$v-1$};
\node[above=0.28cm of v] {$v$};

\node[below=0.32cm of v1] {$a_i^{(1)}$};
\node[below=0.32cm of v2] {$a_i^{(2)}$};
\node[below=0.32cm of v3] {$a_i^{(3)}$};
\node[below=0.32cm of vm1] {$a_i^{(v-1)}$};
\node[below=0.32cm of v] {$a_i^{(v)}$};

\node[
    draw,
    dashed,
    minimum width=1.9cm,
    minimum height=1.0cm,
    align=center
] (target) at (13.55,0)
{$Y_i(a_i^{(v)}+\tau)$};

\draw[->, thick] (v.east) -- node[above, midway] {$\tau$} (target.west);

\node[above=0.28cm of target] {Prediction target};
\node[below=0.32cm of target] {$a_i^{(v)}+\tau$};

\end{tikzpicture}

\caption{Observed longitudinal history with unequal elapsed intervals $\Delta a_i^{(v)}$ and prospective prediction horizon $\tau$.}
\label{fig:timing_structure}
\end{figure*}

Define the observed longitudinal history through visit $v$ as
\begin{equation}
\mathcal{H}_i^{(v)}
=
\left\{
\left(a_i^{(\ell)},\mathbf{x}_i^{(\ell)},
\mathbf{c}_i^{(\ell)}\right):1\leq\ell\leq v
\right\}.
\end{equation}
The observed ages determine the historical elapsed intervals, which are used to represent the timing of prior health observations.

For a prospective prediction horizon $\tau>0$, the model estimates the probability that an individual observed through visit $v$ occupies state $s$ at age $a_i^{(v)}+\tau$, conditional on the observed longitudinal history and the demographic and socioeconomic information available at visit $v$. Let $\theta$ denote the collection of trainable model parameters. The horizon-specific conditional transition distribution is
\begin{equation}
\label{eq:conditional-transition}
p_{\theta,is}^{(v)}(\tau)
=
\Pr_\theta\!\left(
Y_i(a_i^{(v)}+\tau)=s
\mid \mathcal{H}_i^{(v)},\tau
\right),
\qquad s\in\mathcal{J}.
\end{equation}

The prediction horizon $\tau$ is distinct from the historical elapsed interval $\Delta a_i^{(v)}$: $\Delta a_i^{(v)}$ describes spacing within the observed history, whereas $\tau$ specifies the future horizon over which the transition probability is evaluated.

The learned map $f_\theta$ produces the corresponding probability vector
\begin{equation}
\label{eq:model-output}
\widehat{\mathbf{p}}_i^{(v)}(\tau) = f_\theta\!\left(\mathcal{H}_i^{(v)}, \tau
\right) =
\left(
\widehat p_{iH}^{(v)}(\tau),
\widehat p_{iM}^{(v)}(\tau),
\widehat p_{iS}^{(v)}(\tau),
\widehat p_{iD}^{(v)}(\tau)
\right).
\end{equation}
The output satisfies
$$
\widehat p_{is}^{(v)}(\tau)\geq0, \qquad \sum_{s\in\mathcal{J}}\widehat p_{is}^{(v)}(\tau)=1.
$$

Thus, $\widehat p_{is}^{(v)}(\tau)$ is the individual-level predicted probability of occupying destination state $s$ after horizon $\tau$. The model is explicitly conditioned on the prediction horizon, allowing evaluation at supported horizons rather than assuming a single fixed follow-up interval. In the empirical analysis, the two-year horizon is the primary actuarial setting, with additional horizons examined where sufficient longitudinal support is available. These individual-level horizon-specific probabilities are aggregated into actuarial transition matrices in Section~\ref{subsec:aggregation}.

\subsection{\lantern{} Architecture}
\label{subsec:architecture}

\hspace*{-\parindent}\textit{Historical Observation-Time Representation.} The historical elapsed interval $\Delta a_i^{(v)}$ measures the age difference between consecutive observed visits and
is distinct from the prospective prediction horizon $\tau$. To retain information about irregular observation spacing, we construct
\begin{equation}
\widetilde{\Delta a}_i^{(v)}
=
\log\!\left(1+\Delta a_i^{(v)}\right),
\end{equation}
where the elapsed interval is set to zero for the first retained visit. The numerical input at visit $v$ is augmented with the transformed elapsed interval, a first-visit indicator, and cumulative visit count:
\begin{equation}
\label{eq:augmented-input}
\widetilde{\mathbf{x}}_i^{(v)}
=
\left[
\mathbf{x}_i^{(v)},
\widetilde{\Delta a}_i^{(v)},
f_i^{(v)},
n_i^{(v)}
\right],
\end{equation}
where $f_i^{(v)}$ indicates the first retained visit and $n_i^{(v)}$ denotes the number of previously processed visits.
These variables allow the recurrent model to account for historical observation spacing and visit progression. No separate periodic time embedding is used in the final specification. The prospective horizon $\tau$ enters the prediction head after the recurrent history representation has been updated.

\paragraph{\it Adaptive Attribute Conditioning}
The predictive relevance of demographic and socioeconomic attributes to health-state transitions may vary across individuals and over the life course as their health conditions and functional histories evolve. Even when an attribute remains unchanged across visits, its relevance to future health outcomes may depend on the individual's accumulated health history. To accommodate this heterogeneity, \lantern{} employs multi-head attention to construct an attribute representation conditioned on the preceding recurrent memory.

Unlike direct concatenation of attribute inputs, the attention mechanism explicitly conditions their relative weighting on the preceding recurrent memory. This provides a structured way to construct history-dependent attribute representations, although comparable interactions could also be learned by other sufficiently flexible neural architectures.

Let $\mathbf{e}_{ij}^{(v)}\in\mathbb{R}^{d_h}$ denote the learned embedding of attribute $j$ for individual $i$ at visit $v$, and let
$\mathbf{E}_i^{(v)}=[\mathbf{e}_{i1}^{(v)},\ldots,\mathbf{e}_{iA}^{(v)}]^\top
\in\mathbb{R}^{A\times d_h}$
collect the attribute embeddings. The attention query is formed from the preceding recurrent memory $\mathbf{q}_i^{(v)} = h_\theta\!\left(\mathbf{m}_i^{(v-1)}\right),$ where $h_\theta$ is a learned nonlinear projection. For attention head $k=1,\ldots,K$, define
\begin{equation}
    \boldsymbol{\alpha}_{ik}^{(v)} = \operatorname{softmax}\!\left(
    \frac{(\mathbf{q}_i^{(v)}W_k^Q) (\mathbf{E}_i^{(v)}W_k^K)^\top}{\sqrt{d_k}}
    \right),
\end{equation}
where $W_k^Q$ and $W_k^K$ are learned projection matrices and $d_k=d_h/K$ is the dimension of each head. The corresponding head output is $\mathbf{o}_{ik}^{(v)} = \boldsymbol{\alpha}_{ik}^{(v)}(\mathbf{E}_i^{(v)}W_k^V),$ where $W_k^V$ is the learned value projection. The adaptive attribute representation is
\begin{equation}
    \label{eq:attribute-attention}
    \mathbf{u}_i^{(v)} = \operatorname{Concat} \left(
    \mathbf{o}_{i1}^{(v)},\ldots,\mathbf{o}_{iK}^{(v)}
    \right)W^O.
\end{equation}

For diagnostic purposes, attention weights are averaged across the $K$ heads, yielding $\bar{\alpha}_{ij}^{(v)}=K^{-1}\sum_{k=1}^{K}\alpha_{ijk}^{(v)}$. These weights characterize the relative allocation of attention among attributes and are not interpreted as causal effects or independent measures of predictive importance.

Because attention is conditioned on preceding recurrent memory, the weights assigned to an attribute may change across visits even when its observed value remains constant. For time-varying attributes, such as marital status and census region, the weights may additionally respond to changes in their observed values.

\paragraph{\it Memory Update and Transition Modeling}
\label{subsec:modpred}
For each individual $i$, \lantern{} maintains a latent memory vector $\mathbf{m}_i^{(v)}\in\mathbb{R}^{d_h}$ that summarizes the observed health trajectory through visit $v$. The memory is updated sequentially using a gated recurrent unit cell (GRUCell) \cite{cho2014learning}, which combines the augmented numerical features and adaptive attribute
representation with the preceding memory:
\begin{equation}
\label{eq:latent-memory}
    \mathbf{m}_i^{(v)} = \mathrm{GRUCell}_{\theta}\!\left(
    [\widetilde{\mathbf{x}}_i^{(v)},\mathbf{u}_i^{(v)}],
    \mathbf{m}_i^{(v-1)}
    \right),
    \qquad
    \mathbf{m}_i^{(0)}=\mathbf{0}.
\end{equation}

The recurrent update is applied separately at each observed visit using the individual's previously stored memory; the wave-wise training and memory-management procedure is described in Section~\ref{exp}. Figure~\ref{fig:hist_dep} illustrates how the observed health history is summarized through recurrent memory and combined with the prospective prediction horizon to estimate transition probabilities.

\begin{figure}[h]
\centering
\begin{tikzpicture}[
    >=Latex,
    font=\small,
    visit/.style={
        draw,
        rounded corners=2pt,
        minimum width=1.10cm,
        minimum height=0.75cm,
        align=center
    },
    state/.style={
        draw,
        rounded corners=2pt,
        minimum width=1.65cm,
        minimum height=0.90cm,
        align=center
    },
    output/.style={
        draw,
        rounded corners=2pt,
        minimum width=1.75cm,
        minimum height=0.78cm,
        align=center
    },
    arrow/.style={
        ->,
        thick
    }
]

\node[visit] (v1) at (0,0) {$\mathbf{x}_i^{(1)}$};
\node[visit] (v2) at (1.55,0) {$\mathbf{x}_i^{(2)}$};
\node at (3.05,0) {$\cdots$};
\node[visit] (vv) at (4.55,0) {$\mathbf{x}_i^{(v)}$};

\node[state] (mem) at (7.10,0)
{\scriptsize History encoder\\[2pt]$\mathbf{m}_i^{(v)}$};

\node[output] (out) at (9.75,0)
{$\widehat{\mathbf{p}}_i^{(v)}(\tau)$};

\node[visit] (tau) at (9.75,-1.45) {$\tau$};

\draw[arrow] (v2.east) -- (2.65,0);
\draw[arrow] (3.45,0) -- (vv.west);

\draw[arrow] (vv.east) -- (mem.west);

\draw[arrow] (mem.east) -- (out.west);

\draw[arrow] (tau.north) -- (out.south);

\draw[decorate, decoration={brace,mirror,amplitude=4pt}]
(-0.45,-0.58) -- (5.00,-0.58)
node[midway,below=5pt,font=\scriptsize]
{observed health history};

\end{tikzpicture}

\caption{\textbf{History dependence in \lantern{}.}
The observed health trajectory through visit $v$ is summarized by the latent memory $\mathbf{m}_i^{(v)}$. Together with the prospective prediction horizon $\tau$, this history representation is used to estimate the horizon-specific transition probability vector $\widehat{\mathbf{p}}_i^{(v)}(\tau)$.}
\label{fig:hist_dep}
\end{figure}

The updated latent memory $\mathbf{m}_i^{(v)}$ summarizes available trajectory information through visit $v$. To obtain a horizon-specific prediction, the prospective horizon $\tau$ is supplied to the transition head together with the latent memory representation. Thus,
\begin{equation}
\label{eq:horizon-head}
\widehat{\mathbf{p}}_i^{(v)}(\tau)=g_\theta\!\left(\mathbf{m}_i^{(v)},\tau\right),
\end{equation}
where $g_\theta$ denotes the horizon-conditioned transition mapping.

Given $\mathbf{m}_i^{(v)}$ and $\tau$, the model predicts the transition distribution at age $a_i^{(v)}+\tau$ using a hierarchical factorization that separates mortality from transitions among living disability states. Mortality risk is modeled as a Bernoulli component with logit
$$
z_{D,i}^{(v)}(\tau) = g_D\!\left(\mathbf{m}_i^{(v)},\tau\right).
$$
Hence, the predicted probability of death is
\begin{equation}
\widehat p_{iD}^{(v)}(\tau) = \sigma\!\left(z_{D,i}^{(v)}(\tau)\right),
\end{equation}
where $\sigma(\cdot)$ denotes the sigmoid function.

Let $\mathcal{J}_{\rm alive}=\{H,M,S\}\subset\mathcal{J}$ denote the set of non-death states. Conditional on survival, the transition distribution over $\mathcal{J}_{\rm alive}$ is modeled using a multinomial component with logit vector
$$
\mathbf{z}_{{\rm alive},i}^{(v)}(\tau) = g_{\rm alive}\!\left(\mathbf{m}_i^{(v)},\tau\right)\in\mathbb{R}^{3}.
$$
The conditional probability of alive state $s\in\mathcal{J}_{\rm alive}$ is
$$
\rho_{is}^{(v)}(\tau) = \frac{\exp\!\left(z_{{\rm alive},is}^{(v)}(\tau)\right)}{\sum_{s'\in\mathcal{J}_{\rm alive}}\exp\!\left(z_{{\rm alive},is'}^{(v)}(\tau)\right)}, \qquad s\in\mathcal{J}_{\rm alive}.
$$
Together with the predicted death probability $\widehat p_{iD}^{(v)}(\tau)$, the living-state probabilities are
\begin{equation}
\widehat p_{is}^{(v)}(\tau) = \left(1-\widehat p_{iD}^{(v)}(\tau)\right)\rho_{is}^{(v)}(\tau), \qquad s\in\mathcal{J}_{\rm alive}.
\end{equation}
By construction,
$$
\sum_{s\in\mathcal{J}} \widehat p_{is}^{(v)}(\tau) = 1 \quad \text{and}\quad \widehat p_{is}^{(v)}(\tau)\geq 0
$$
for every individual-history and prediction-horizon combination.

The resulting vector $\widehat{\mathbf{p}}_i^{(v)}(\tau)$ is therefore the predicted health-state distribution at age $a_i^{(v)}+\tau$. Death is enforced as absorbing when the individual-level probabilities are aggregated into actuarial projection matrices.

\subsection{Aggregation into Actuarial Transition Matrices}
\label{subsec:aggregation}

The output of our model is an individual-level, horizon-specific transition probability vector $\widehat{\mathbf{p}}_i^{(v)}(\tau)$. For actuarial analysis, these individual-level probabilities are aggregated into age- and origin-state-specific transition matrices at a common forecast horizon $\tau$. Let $\mathcal{I}_{ag,r}(\tau)$ denote the set of test-set individual-visit observations with current age in age group $ag$, current observed state $r$, and realized forecast horizon $\tau$:
$$\mathcal{I}_{ag,r}(\tau) = \left\{(i,v)\in\mathcal{D}_{\mathrm{test}}(\tau): a_i^{(v)}\in ag,\; Y_i^{(v)}=r \right\}$$

For each living origin state $r\in\{H,M,S\}$ and destination state $s\in\mathcal{J}$, the aggregated transition probability is
\begin{equation}
\label{eq:aggregated-transition}
\widehat p_{rs}(ag,\tau)
=
\frac{1}{|\mathcal{I}_{ag,r}(\tau)|}
\sum_{(i,v)\in\mathcal{I}_{ag,r}(\tau)}
\widehat p_{is}^{(v)}(\tau).
\end{equation}

Here, $\widehat p_{rs}(ag,\tau)$ denotes the estimated probability of transitioning from origin state $r$ to destination state $s$ over horizon $\tau$, averaged across individuals in the corresponding age-origin-state group. The resulting transition matrix is
$$\widehat{\mathbf{P}}(ag,\tau) = \left(\widehat p_{rs}(ag,\tau)\right)_{r,s\in\mathcal{J}}$$

Death is treated as an absorbing state by setting 
$$\widehat p_{DD}(ag,\tau)=1, \qquad \widehat p_{Ds}(ag,\tau)=0 \quad\text{for }s\neq D.$$

By construction, the resulting transition matrix is row-stochastic and retains the four-state actuarial structure. Its living-state transition probabilities aggregate history-conditioned individual predictions within each age-origin-state group at a specified horizon. Although these matrices can be evaluated against observed transition frequencies, their repeated use in long-term cohort projections requires additional assumptions about the evolution of individual health histories and covariates.

\section{Numerical Experiments}
\label{exp}

\subsection{Training Setup}

The hierarchical prediction structure motivates a conditional loss that distinguishes mortality from transitions among living health states. The mortality head is trained on all observed transitions, while the living-state head is trained only on transitions ending in Healthy, Mild, or Severe states, since a living-state outcome is not defined following death.

Let $\mathcal{B}$ denote a training mini-batch and $\mathcal{B}_{\mathrm{alive}}\subseteq\mathcal{B}$ the subset with living destination states. For transition $(i,v)$, define $y_{D,i}^{(v)}=\mathbf{1}\{Y_i^{(v+1)}=D\}$, and let $\tau_i^{(v)}$ denote the observed follow-up interval. The mini-batch loss is
\begin{equation}
\label{eq:training-loss}
\mathcal{L}_{\mathcal{B}}(\theta)
=
\frac{1}{|\mathcal{B}|}
\sum_{(i,v)\in\mathcal{B}}
\ell_{\mathrm{BCE}}\!\left(
y_{D,i}^{(v)},
z_{D,i}^{(v)}(\tau_i^{(v)})
\right)
+
\frac{1}{|\mathcal{B}_{\mathrm{alive}}|}
\sum_{(i,v)\in\mathcal{B}_{\mathrm{alive}}}
\ell_{\mathrm{CE}}\!\left(
Y_i^{(v+1)},
\mathbf{z}_{{\rm alive},i}^{(v)}(\tau_i^{(v)})
\right)
\end{equation}
The two components are averaged separately and combined without class weighting. When a mini-batch contains no living destination states, the second component is omitted.

Data are partitioned by individual into training (70\%), validation (15\%), and test (15\%) sets to prevent cross-partition leakage. Observations retain their temporal ordering, and predictors are restricted to information available through the current visit. The observed follow-up interval is supplied as the prediction horizon during training.

\paragraph{\it Wave-Wise Recurrent Training}
Unlike conventional recurrent training based on padded individual-level sequences, \lantern{} processes observations in chronological survey-wave batches. At each wave, the model retrieves the stored latent memory for each observed individual and updates it using the current numerical features, historical elapsed-time information, and attention-conditioned attribute representation. The updated memory is retained for subsequent observations of the same individual, while individuals absent from a given wave preserve their previous memory state. This procedure accommodates unequal numbers of observations and irregular intervals between visits without requiring fixed-length individual sequences.

Individual-specific memories are initialized to zero at the beginning of each epoch and updated sequentially across observed waves. Model parameters are optimized after each wave batch, with gradients restricted to the current recurrent update rather than propagated through complete individual histories. Validation and test inference follow the same chronological memory-update procedure using separately initialized memories.

The model uses a recurrent hidden dimension of 128 and four attribute-attention heads. Optimization employs Adam with a learning rate of $3\times10^{-3}$, weight decay of $10^{-6}$, and gradient clipping at a norm of 1.0. Training is limited to 50 epochs, with early stopping based on validation loss and a fixed random seed of 42. Model selection was performed by retaining the checkpoint with the lowest validation loss rather than conducting a systematic hyperparameter search.

\subsection{Evaluation Metrics}

Predictive performance is evaluated on held-out test observations at their realized forecast horizons. For $\tau\in\{1,2,3\}$, let $\mathcal{D}_{\mathrm{test}}(\tau)$ denote the subset of test transitions with realized horizon $\tau$. Paired comparisons between \lantern{} and the benchmarks use 1,000 individual-cluster bootstrap replicates to account for repeated observations within respondents.

\paragraph{\it Multi-State Probabilistic Accuracy}
Overall predictive accuracy is evaluated using the multiclass Brier score:
\begin{equation}
    \mathrm{Brier}_{\mathrm{MC}}(\tau) =
    \frac{1}{|\mathcal{D}_{\mathrm{test}}(\tau)|}
    \sum_{(i,v)\in\mathcal{D}_{\mathrm{test}}(\tau)}
    \sum_{s\in\mathcal{J}}
    \left(
    \widehat p_{is}^{(v)}(\tau)
    -
    \mathbf{1}\{Y_i^{(v+1)}=s\}
    \right)^2.
\end{equation}
This proper scoring rule measures the agreement between the predicted four-state probability distribution and the observed destination state, with lower values indicating greater accuracy.

\paragraph{\it Calibration}
Calibration is assessed using reliability diagrams and expected calibration error (ECE). For severe disability and mortality, predicted endpoint probabilities are grouped into probability intervals, and observed event frequencies are compared with mean predicted probabilities. ECE summarizes the absolute differences between observed and predicted probabilities, weighted by the number of observations in each interval. Multiclass confidence calibration is considered separately from endpoint-specific calibration.

\paragraph{\it Binary Endpoint Discrimination}
Severe disability and mortality are evaluated separately using the corresponding marginal probabilities from the four-state prediction vector. For each endpoint, we report the area under the receiver operating characteristic curve (AUROC) and average precision (AP). These measures assess discrimination and the precision--recall trade-off, respectively, and complement the probabilistic accuracy and calibration analyses.

\paragraph{\it Fixed-Horizon Transition-Matrix Accuracy}
Model-implied transition probabilities are aggregated by age group and living origin state, as described in Section~\ref{subsec:aggregation}, and compared with empirical transition frequencies at the same horizon. For each age group $ag$, origin state $r$, and destination state $s$, the empirical transition probability is
\begin{equation}
    \widetilde p_{rs}(ag,\tau)
    =
    \frac{
    \sum_{(i,v)\in\mathcal{I}_{ag,r}(\tau)}
    \mathbf{1}\{Y_i^{(v+1)}=s\}
    }{
    |\mathcal{I}_{ag,r}(\tau)|
    }.
\end{equation}
Let $\mathcal{K}_{\tau}$ denote the set of evaluated age-group, living-origin-state, and destination-state cells. Transition-matrix accuracy is summarized using
\begin{equation}
    \mathrm{MAE}_{P}(\tau)
    =
    \frac{1}{|\mathcal{K}_{\tau}|}
    \sum_{(ag,r,s)\in\mathcal{K}_{\tau}}
    \left|
    \widehat p_{rs}(ag,\tau)
    -
    \widetilde p_{rs}(ag,\tau)
    \right|
\end{equation}
and
\begin{equation}
    \mathrm{RMSE}_{P}(\tau)
    =
    \left[
    \frac{1}{|\mathcal{K}_{\tau}|}
    \sum_{(ag,r,s)\in\mathcal{K}_{\tau}}
    \left(
    \widehat p_{rs}(ag,\tau)
    -
    \widetilde p_{rs}(ag,\tau)
    \right)^2
    \right]^{1/2}
\end{equation}
We also compute a transition-count-weighted MAE:
\begin{equation}
\label{eq:weighted-matrix-mae}
    \mathrm{WMAE}_{P}(\tau) = \frac{\sum_{(ag,r)\in\mathcal{G}_{\tau}} n_{ag,r}(\tau) \left[\frac{1}{4}\sum_{s\in\mathcal{J}}\left| \widehat p_{rs}(ag,\tau)
    -
    \widetilde p_{rs}(ag,\tau) \right| \right]}{\sum_{(ag,r)\in\mathcal{G}_{\tau}} n_{ag,r}(\tau)}
\end{equation}
Here, $n_{ag,r}(\tau)=|\mathcal{I}_{ag,r}(\tau)|$,
and $\mathcal{G}_{\tau}$ contains the evaluated age-origin-state groups. The analysis considers four age groups, three living origin states, and four destination states. For the paired individual-cluster bootstrap, groups with fewer than 30 observed transitions are excluded, and eligible groups are fixed before resampling.

\subsection{Baselines}

We compare \lantern{} with four benchmarks. Multinomial Logistic Regression \cite{hastie2009elements} and LightGBM \cite{ke2017lightgbm} estimate four-state transition probabilities from tabular covariates, including the forecast horizon. The logistic model uses L2 regularization, while LightGBM captures nonlinear covariate relationships through gradient-boosted trees.

The horizon-conditioned GRU benchmark \cite{cho2014learning} processes padded individual-level observation sequences using a standard GRU layer trained through sequence-based backpropagation.
The forecast horizon is appended to the recurrent representation before a four-class prediction head. Unlike \lantern{}, the GRU benchmark processes complete individual observation sequences during training and does not employ adaptive attribute attention or the hierarchical mortality and conditional living-state prediction structure.

The actuarial continuous-time Markov chain (CTMC) benchmark models transition intensities as log-linear functions of age, age squared, and sex, with Death treated as absorbing. Transition probabilities are computed as $\mathbf{P}(\tau)=\exp\{\tau\mathbf{Q}(a+\tau/2)\}$, where $\mathbf{Q}$ is the fitted intensity matrix evaluated at the midpoint age of the observation interval. The CTMC parameters are estimated by maximizing the likelihood of the observed panel transitions.

All models use the same individual-level training, validation, and test partitions. The machine learning benchmarks use the same underlying observed health, demographic, and socioeconomic
variables as \lantern{}, but differ in how they represent longitudinal history. The actuarial CTMC serves as a restricted age- and sex-dependent reference model.

\section{Empirical Results}
\label{sec:results}

\subsection{Predictive Performance Across Forecast Horizons}
\label{subsec:predictive_performance}
Table~\ref{tab:predictive_performance} summarizes the predictive performance of LANTERN and the benchmark models across the one-, two-, and three-year forecast horizons.

\begin{table}[h]
\centering
\caption{Predictive performance across forecast horizons. Lower Brier scores and higher AUROC and AP values indicate better performance.}
\label{tab:predictive_performance}
\small
\begin{tabular}{llccccc}
\hline
\textbf{Horizon} & \textbf{Model} & \textbf{Brier}
& \multicolumn{2}{c}{\textbf{Severe disability}}
& \multicolumn{2}{c}{\textbf{Mortality}} \\
\cline{4-7}
& & & \textbf{AUROC} & \textbf{AP} & \textbf{AUROC} & \textbf{AP} \\
\hline
\multirow{5}{*}{1 year}
& LANTERN        & 0.2529 & 0.9238 & 0.6648 & 0.8782 & 0.0261 \\
& GRU            & 0.2491 & 0.9253 & 0.6809 & 0.8877 & 0.0303 \\
& Logistic       & 0.2561 & 0.9166 & 0.6702 & 0.8891 & 0.0503 \\
& LightGBM       & 0.2512 & 0.9218 & 0.6792 & 0.8638 & 0.0235 \\
& Actuarial CTMC & 0.2924 & 0.8498 & 0.5606 & 0.8218 & 0.0282 \\
\hline
\multirow{5}{*}{2 years}
& LANTERN        & 0.2857 & 0.9156 & 0.6484 & 0.8208 & 0.0693 \\
& GRU            & 0.2856 & 0.9151 & 0.6508 & 0.8214 & 0.0626 \\
& Logistic       & 0.2943 & 0.9102 & 0.6409 & 0.8256 & 0.0682 \\
& LightGBM       & 0.2881 & 0.9127 & 0.6422 & 0.8196 & 0.0577 \\
& Actuarial CTMC & 0.3132 & 0.8750 & 0.5418 & 0.7990 & 0.0608 \\
\hline
\multirow{5}{*}{3 years}
& LANTERN        & 0.3160 & 0.9000 & 0.5864 & 0.8091 & 0.1391 \\
& GRU            & 0.3180 & 0.8964 & 0.5762 & 0.8123 & 0.1327 \\
& Logistic       & 0.3267 & 0.8921 & 0.5762 & 0.8134 & 0.1416 \\
& LightGBM       & 0.3206 & 0.8963 & 0.5741 & 0.8084 & 0.1187 \\
& Actuarial CTMC & 0.3485 & 0.8488 & 0.4786 & 0.7880 & 0.1259 \\
\hline
\end{tabular}

\vspace{2mm}
\begin{minipage}{0.97\linewidth}
\footnotesize
\textit{Note:} The one-, two-, and three-year test samples contain 4,261, 24,419, and 4,253 transitions, respectively, including 21, 299, and 118 observed deaths. Corresponding severe-disability endpoints number 457, 2,689, and 493. AP denotes average precision.
\end{minipage}
\end{table}

At the primary two-year horizon, LANTERN achieves a multiclass Brier score of 0.2857, comparable to GRU (0.2856) and lower than LightGBM (0.2881), logistic regression (0.2943), and the actuarial CTMC (0.3132). Its difference relative to LightGBM is supported by the paired individual-cluster bootstrap analysis ($-0.0024$; 95\% CI: $[-0.0037,-0.0011]$). For Severe disability, our model achieves an AUROC of 0.9156 and average precision of 0.6484, performing comparably to the strongest machine-learning benchmarks and exceeding the CTMC (AUROC: 0.8750; AP: 0.5418).

Mortality prediction is more challenging because of the relatively small number of observed deaths. At two years, LANTERN achieves a mortality AUROC of 0.8208 and the highest average precision among the evaluated models (0.0693). Its average-precision difference relative to LightGBM is 0.0116 (95\% CI: [0.0036, 0.0220]). Nevertheless, mortality AUROC values are similar across the machine-learning models, and absolute average precision remains low.

At the one-year horizon, GRU achieves the lowest Brier score (0.2491), whereas LANTERN achieves the lowest Brier score at three years (0.3160). Our model also obtains the highest Severe-disability AUROC and average precision at three years. However, the relative performance of the models varies across horizons and endpoints. In particular, the one-year mortality results warrant caution because only 21 deaths are observed in that evaluation subset.

\subsection{Calibration of Severe-Disability and Mortality Probabilities}
\label{subsec:calibration}
\begin{figure}[h]
\centering
\includegraphics[width=0.8\linewidth]{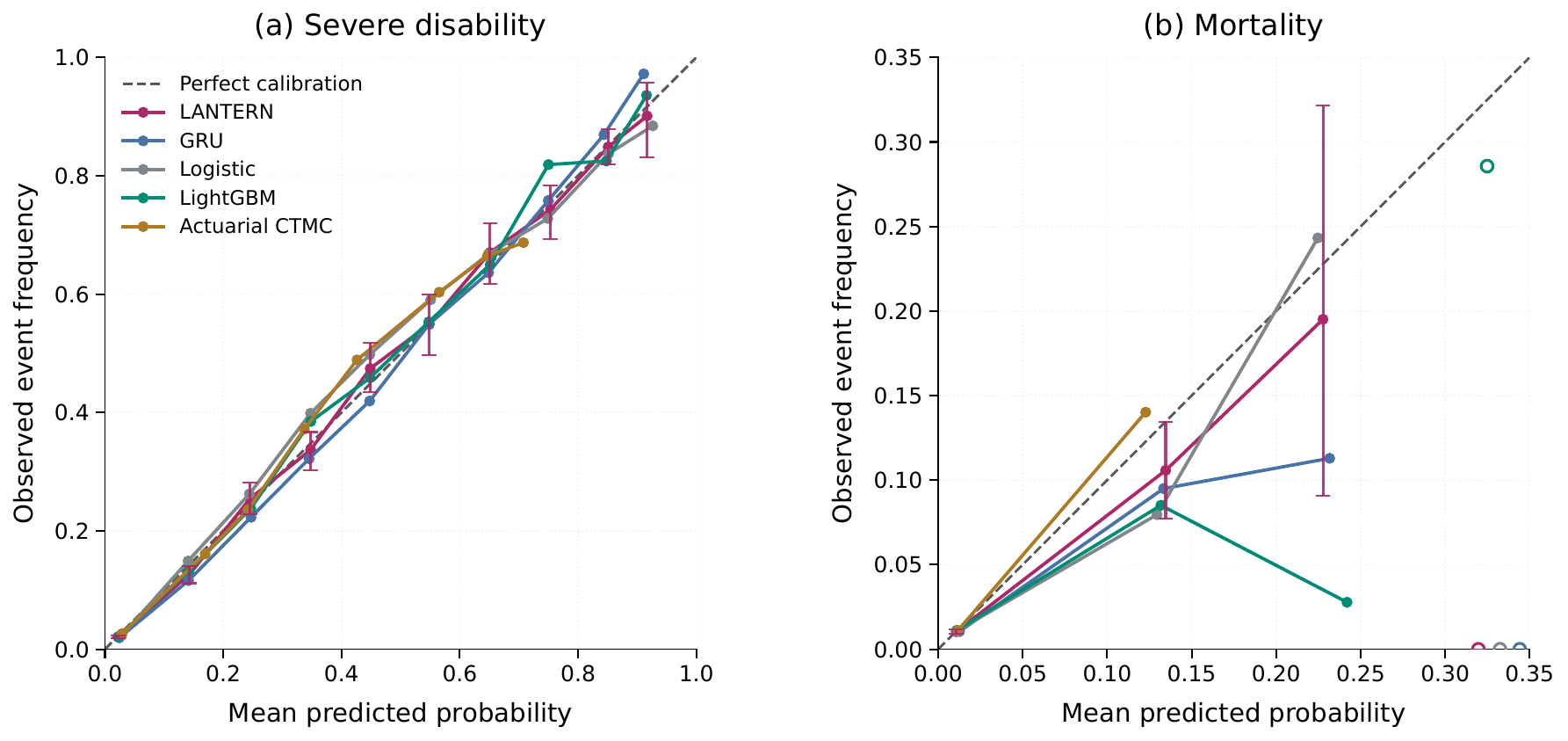}
\caption{Calibration of predicted probabilities of Severe disability (a) and mortality (b) at the two-year forecast horizon. Predicted probabilities are compared with observed event frequencies within probability intervals for LANTERN and the benchmark models. The diagonal represents perfect calibration, and vertical error bars indicate 95\% individual-cluster bootstrap confidence intervals for LANTERN.}
\label{fig:calibration_tau2}
\end{figure}

Figure~\ref{fig:calibration_tau2} presents the calibration of predicted Severe-disability and mortality probabilities at the primary two-year horizon. For Severe disability, LANTERN achieves an expected calibration error (ECE) of 0.0043, compared with 0.0053 for LightGBM, 0.0059 for the actuarial CTMC, 0.0094 for GRU, and 0.0096 for logistic regression. The calibration curve generally follows the diagonal, indicating agreement between predicted probabilities and observed event frequencies across much of the evaluated probability range, although deviations occur in some intervals.

For mortality, LANTERN achieves an ECE of 0.0008, comparable to the actuarial CTMC (0.0007) and lower than those of logistic regression (0.0010), LightGBM (0.0013), and GRU (0.0031). However, predicted mortality probabilities are concentrated near zero. Of the 24,419 two-year test observations, 23,978 fall within the lowest mortality-risk interval, while only 44 have predicted probabilities above 0.20. Consequently, the low aggregate ECE primarily reflects the well-populated low-risk region and does not establish reliable calibration at higher predicted mortality risks, where empirical support is limited.

The one- and three-year calibration results are provided in Supplementary Figure~\ref{fig:calibration_all_horizons}. These additional assessments should be interpreted in light of their smaller evaluation samples, particularly the limited number of observed deaths at one year.

\subsection{Accuracy of Health-State Transition Probability Estimates}
\label{subsec:transition_accuracy}
Table~\ref{tab:transition_accuracy} reports the accuracy of estimated transition probabilities across the one-, two-, and three-year horizons. Mean absolute error (MAE) and root mean squared error (RMSE) compare observed and predicted transition probabilities across 48 age-specific matrix cells, corresponding to four age groups, three living origin states, and four possible destination states.

\begin{table}[h]
\centering
\caption{Accuracy of estimated health-state transition probabilities across forecast horizons. Lower MAE and RMSE indicate closer agreement with observed transition probabilities. Weighted MAE accounts for the number of observations contributing to each origin-state group.}
\label{tab:transition_accuracy}
\small
\begin{tabular}{llccc}
\hline
\textbf{Horizon} & \textbf{Model} & \textbf{MAE} & \textbf{RMSE} & \textbf{Weighted MAE} \\
\hline
\multirow{5}{*}{1 year}
& LANTERN        & 0.0231 & 0.0321 & 0.0142 \\
& GRU            & 0.0229 & 0.0327 & 0.0126 \\
& Logistic       & 0.0304 & 0.0428 & 0.0195 \\
& LightGBM       & 0.0196 & 0.0285 & 0.0115 \\
& Actuarial CTMC & 0.0597 & 0.0885 & 0.0337 \\
\hline
\multirow{5}{*}{2 years}
& LANTERN        & 0.0097 & 0.0141 & 0.0067 \\
& GRU            & 0.0117 & 0.0157 & 0.0077 \\
& Logistic       & 0.0236 & 0.0365 & 0.0150 \\
& LightGBM       & 0.0095 & 0.0139 & 0.0056 \\
& Actuarial CTMC & 0.0097 & 0.0142 & 0.0063 \\
\hline
\multirow{5}{*}{3 years}
& LANTERN        & 0.0270 & 0.0350 & 0.0148 \\
& GRU            & 0.0320 & 0.0458 & 0.0162 \\
& Logistic       & 0.0303 & 0.0416 & 0.0184 \\
& LightGBM       & 0.0220 & 0.0298 & 0.0124 \\
& Actuarial CTMC & 0.0374 & 0.0511 & 0.0209 \\
\hline
\end{tabular}
\end{table}

At the primary two-year horizon, LANTERN achieves a transition-matrix MAE of 0.0097 and RMSE of 0.0141. Its MAE is lower than those of GRU (0.0117) and logistic regression (0.0236), and nearly identical to those of LightGBM (0.0095) and the actuarial CTMC (0.0097). These results indicate close agreement between our model's aggregated predictions and observed transition frequencies, although the strongest benchmarks achieve comparable aggregate accuracy.

\begin{figure}[h]
\centering
\includegraphics[width=\linewidth]{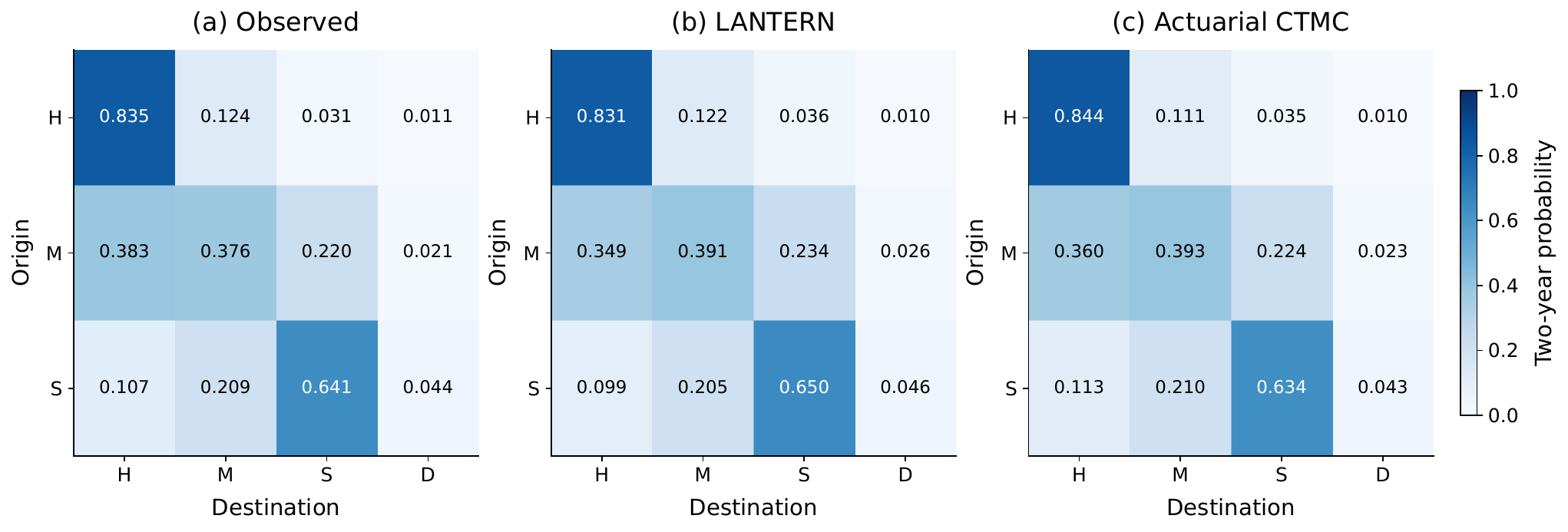}
\caption{Observed and estimated two-year health-state transition matrices for individuals aged 70--79. The panels display observed transition probabilities, LANTERN estimates, and actuarial CTMC estimates. Rows correspond to the living origin states (Healthy, Mild disability, and Severe disability), while columns represent the destination states (Healthy, Mild disability, Severe disability, and Death). Death is an absorbing state and is therefore excluded as an origin state.}
\label{fig:transition_matrices_tau2}
\end{figure}

Figure~\ref{fig:transition_matrices_tau2} illustrates the observed and estimated two-year transition matrices for individuals aged 70--79. Both LANTERN and the actuarial CTMC reproduce the principal features of the observed transition structure, including relatively high probabilities of remaining Healthy or Severely disabled. For example, the observed probability of remaining Healthy is 0.835, compared with 0.831 for LANTERN and 0.844 for the CTMC. The observed probability of transitioning from Mild disability to Healthy is 0.383, whereas our model and the CTMC estimate 0.349 and 0.360, respectively. These differences illustrate that close aggregate matrix accuracy does not imply equally accurate estimation of every transition. The age-specific matrix is presented as an illustration, while Table~\ref{tab:transition_accuracy} summarizes performance across all evaluated age groups.

At one and three years, LANTERN achieves matrix MAEs of 0.0231 and 0.0270, respectively. LightGBM produces the lowest MAE at both horizons (0.0196 and 0.0220), while the actuarial CTMC exhibits its largest error at one year (0.0597). Because the horizon-specific evaluation subsets differ in size and composition, these differences should not be attributed solely to forecast duration.

\subsection{Age-Specific Health-State Transition Patterns}
\label{subsec:age_transition_patterns}
\begin{figure}[h]
\centering
\includegraphics[width=\linewidth]{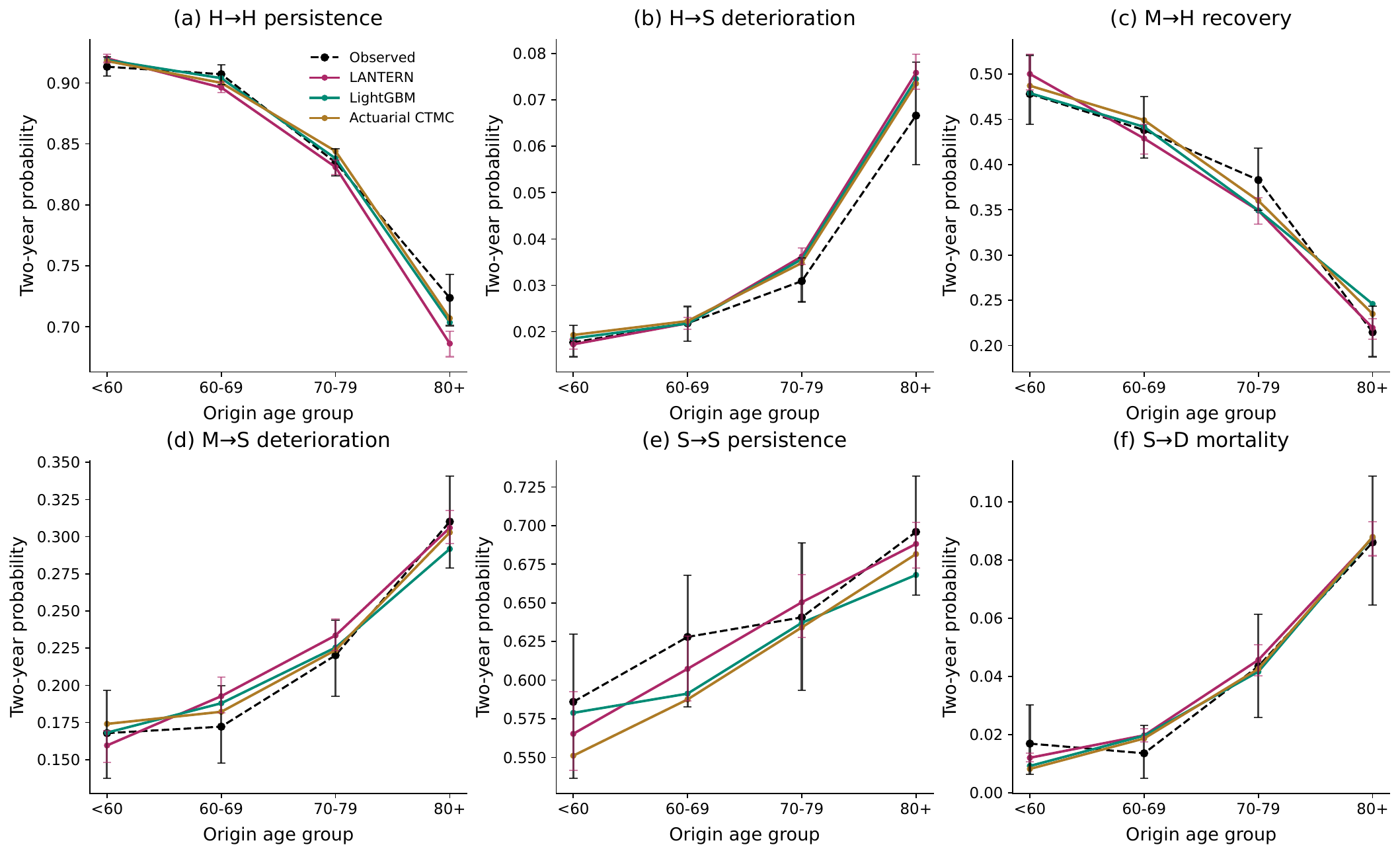}
\caption{Observed and estimated two-year transition probabilities across age groups for six selected health-state transition patterns: (a) Healthy-state persistence, (b) deterioration from Healthy to Severe disability, (c) recovery from Mild disability to Healthy, (d) deterioration from Mild to Severe disability, (e) Severe-disability persistence, and (f) mortality from Severe disability. Predictions from LANTERN, LightGBM, and the actuarial CTMC are compared with observed transition frequencies. Error bars on observed probabilities indicate uncertainty in the empirical estimates.}
\label{fig:age_transition_patterns}
\end{figure}

Figure~\ref{fig:age_transition_patterns} compares observed and estimated two-year transition probabilities across four age groups for selected patterns of persistence, recovery, deterioration, and mortality. The observed probabilities exhibit age-related variation in functional health transitions, which is broadly reproduced by our model and the benchmark models.

The observed probability of remaining Healthy declines from 0.913 among individuals younger than 60 to 0.724 among those aged 80 and older. Correspondingly, the probability of deteriorating from Healthy to either Mild or Severe disability increases from 0.085 to 0.256. LANTERN estimates Healthy-state persistence probabilities of 0.920 and 0.686 and deterioration probabilities of 0.078 and 0.289 for these respective groups. Thus, although the model captures the overall age-related pattern, it underestimates Healthy-state persistence and overestimates deterioration among individuals aged 80 and older.

Recovery from Mild disability to Healthy also declines with age, from an observed probability of 0.478 among individuals younger than 60 to 0.215 among those aged 80 and older. Our model estimates corresponding probabilities of 0.500 and 0.219, although its estimate for ages 70--79 (0.349) is below the observed probability of 0.383. In contrast, deterioration from Mild to Severe disability increases with age, a pattern generally reproduced by the models.

For individuals initially classified as Severely disabled, observed state persistence increases from 0.586 among those younger than 60 to 0.696 among those aged 80 and older. LANTERN estimates probabilities of 0.565 and 0.688 for these groups. Mortality following Severe disability also increases with age, from 0.017 to 0.086, compared with LANTERN estimates of 0.012 and 0.088. Although the general mortality pattern is reproduced, individual estimates should be interpreted cautiously where observed death counts are small.

These findings indicate that our model broadly captures age-related differences in persistence, recovery, deterioration, and mortality. Nevertheless, estimation errors vary across specific transitions and age groups, emphasizing the value of examining individual transition patterns alongside aggregate matrix accuracy.

\subsection{Bootstrap Uncertainty in Transition-Matrix Comparisons}
\label{subsec:transition_uncertainty}
\begin{figure}[h]
\centering
\includegraphics[width=\linewidth]{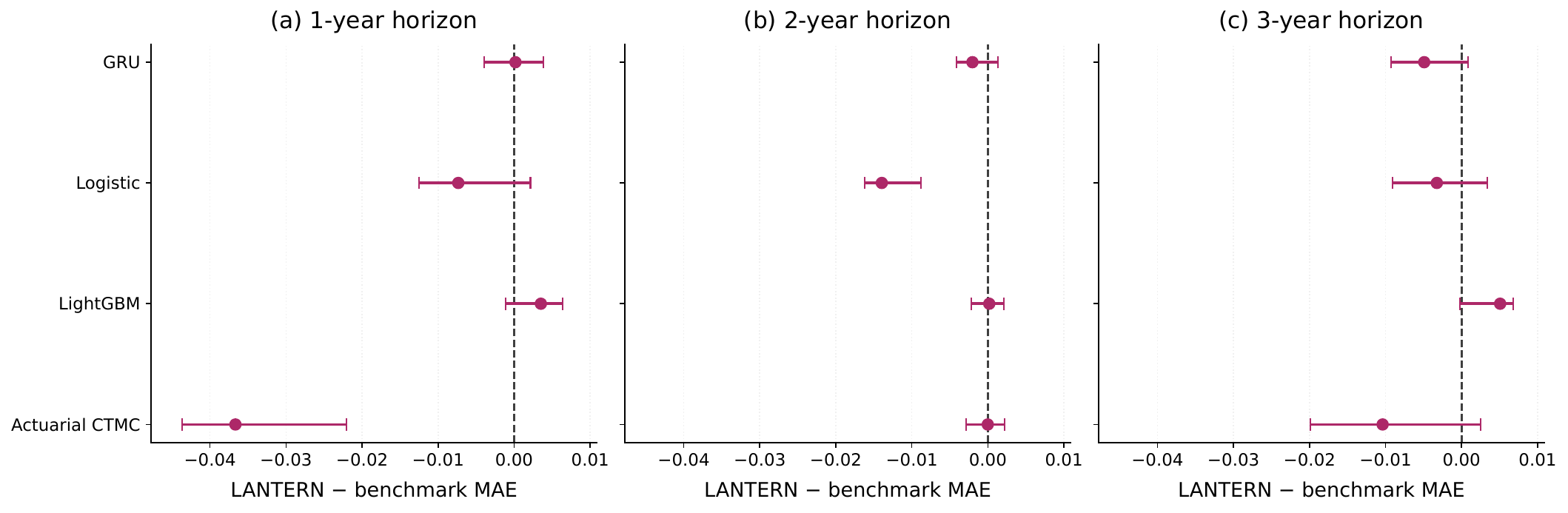}
\caption{Paired differences in transition-matrix mean absolute error (MAE) between LANTERN and the benchmark models at the one-, two-, and three-year forecast horizons. Points represent observed differences (LANTERN minus benchmark), and horizontal intervals represent 95\% individual-cluster bootstrap confidence intervals based on 1,000 bootstrap replicates. Negative differences favor LANTERN. The vertical reference line indicates zero difference.}
\label{fig:matrix_bootstrap}
\end{figure}

Figure~\ref{fig:matrix_bootstrap} presents paired differences in transition-matrix MAE between our model and each benchmark across the three forecast horizons, with 95\% individual-cluster bootstrap confidence intervals. Differences are calculated as LANTERN minus the corresponding benchmark; therefore, negative values favor our model. Intervals spanning zero indicate that the differences are not clearly distinguishable from zero under the bootstrap procedure.

At the primary two-year horizon, our model achieves a lower matrix MAE than logistic regression, with a paired difference of $-0.0139$ (95\% CI: $[-0.0162,-0.0088]$). Its differences relative to GRU ($-0.0020$), LightGBM ($0.0002$), and the actuarial CTMC (approximately $0.0000$) are smaller, and their confidence intervals include zero. Thus, the two-year comparisons provide clearer evidence of improved matrix accuracy relative to logistic regression than relative to the other benchmarks.

At one year, LANTERN achieves a lower matrix MAE than the actuarial CTMC, with a difference of $-0.0367$ (95\% CI: $[-0.0437,-0.0220]$). The intervals for the remaining one-year comparisons include zero. At three years, all four paired confidence intervals span zero, despite differences in the point estimates. The bootstrap results therefore indicate that the strength of the evidence for differences in aggregate transition-matrix accuracy depends on both the benchmark and forecast horizon.

\subsection{Attribute-Attention Diagnostics}
\label{subsec:attention_diagnostics}
At the two-year horizon, learned attribute-attention weights differ between observations in the highest and lowest predicted-risk deciles for Severe disability and mortality (Supplementary Figure~\ref{fig:attention_shift_tau2}). Census region and marital status receive greater average attention in the high-risk groups for both endpoints, while sex exhibits a larger positive difference for mortality. Race and education receive lower relative attention in the high-risk groups. Several differences have 95\% individual-cluster bootstrap confidence intervals excluding zero.

\section{Discussion}
\label{sec:discussion}
\subsection{Longitudinal History and Horizon-Specific Transition Modeling}
The findings support the feasibility of estimating horizon-specific health-state transition probabilities using accumulated individual health histories. This formulation is relevant to disability modeling because individuals occupying the same current health state may differ in their previous functional trajectories, chronic disease histories, and demographic characteristics, potentially resulting in heterogeneous future transition risks. Unlike conventional current-state-dependent Markov formulations, LANTERN accommodates this heterogeneity through history-conditioned predictions without requiring the transition probability to depend exclusively on the current health state. Evaluation across one-, two-, and three-year horizons demonstrates the framework's applicability to the observed HRS follow-up intervals. Empirical support is strongest at two years, reflecting the predominantly biennial survey schedule. The results do not, however, establish predictive validity at arbitrary horizons or isolate the independent contribution of accumulated history.

\subsection{Predictive Performance and Actuarial Implications}
Our model achieves competitive predictive performance relative to conventional actuarial and machine-learning benchmarks, although its advantages are neither uniform nor consistently large. Its overall probabilistic accuracy is comparable to GRU at the primary two-year horizon, while LightGBM achieves similar aggregate transition-matrix accuracy. These findings suggest that the contribution of LANTERN is not universal predictive superiority, but a unified framework for incorporating longitudinal histories, demographic and socioeconomic attributes, and forecast-horizon information into four-state transition probability estimation.

An important distinction emerges between individual-level predictive performance and aggregate transition-matrix accuracy. At the two-year horizon, the actuarial CTMC achieves aggregate matrix accuracy comparable to LANTERN despite weaker performance on several individual-level predictive measures. This difference indicates that models with similar aggregate transition estimates may vary in their ability to distinguish risks across individuals. The choice of modeling approach therefore depends partly on whether the actuarial objective emphasizes individualized risk assessment, group-level transition estimation, or both.

Our model's ability to produce individual predictions that aggregate into plausible age-specific transition probabilities provides a basis for further investigation of such applications. However, its history-conditioned probabilities are not directly interchangeable with a conventional Markov transition system. In particular, repeatedly multiplying matrices obtained by averaging individual predictions would generally fail to account for changes in the distribution of health histories and demographic characteristics over time. Long-term projections of disability occupancy, expected benefit payments, or insurance liabilities would therefore require a procedure that updates individual histories or explicitly models the evolving population distribution. The present evaluation establishes horizon-specific predictive performance rather than the validity of repeated long-term actuarial projections.

\subsection{Health-State Dynamics and Attribute Conditioning}
The age-specific analyses show that \lantern{} broadly reproduces observed patterns of declining functional recovery and increasing disability deterioration and mortality with age. These patterns are relevant to long-term care insurance because movements among functional health states affect the occurrence, persistence, and potential duration of disability. At the same time, differences between observed and estimated probabilities for particular transitions demonstrate that aggregate accuracy alone is insufficient to characterize the model's performance across the health-state system.

The attribute-attention mechanism allows the relative weighting of demographic and socioeconomic representations to vary with an individual's accumulated health history, including for attributes whose observed values remain unchanged. The relevance of these characteristics to later-life health is supported by studies documenting regional variation in functional limitations \cite{porell2002regional}, associations between marital status and mortality \cite{wang2023marital}, and the predictive relevance of sex in HRS-based mortality models \cite{lee2006development}. The observed attention differences across predicted-risk groups provide a limited diagnostic of the model's internal attribute allocation, rather than evidence that these associations are captured by the attention weights. Moreover, attention weights do not establish independent predictive contributions or causal effects, and their interpretation as measures of feature importance remains debated \cite{jain2019attention,wiegreffe2019attention}.

\subsection{Limitations}
Several limitations qualify the findings. First, the predominantly biennial HRS observation structure provides substantially greater empirical support for two-year transitions than for other forecast horizons. The smaller samples at one- and three-year horizons, together with relatively infrequent mortality events, limit the precision and comparability of some predictive and calibration estimates.

Second, health-state classification relies on reported ADL and IADL limitations, while excluding observations with missing functional-status information may affect the composition of the analytic sample. In addition, the evaluation does not isolate the independent contributions of recurrent history and adaptive attribute conditioning, and wave-wise training restricts gradient propagation to individual recurrent updates rather than complete observation histories.

Finally, the empirical evaluation focuses on observed horizon-specific transitions rather than long-term cohort projections. Because individual predictions depend on accumulated health histories and covariates, accurate fixed-horizon transition matrices do not necessarily remain valid when repeatedly applied to cohorts whose characteristics evolve over time.

\section{Conclusion}
\label{sec:conclusion}
This study addresses a central challenge in actuarial health-state modeling: how to estimate transition probabilities at specified future horizons while accounting for accumulated individual health histories, observation timing, and demographic and socioeconomic heterogeneity. We developed \lantern{}, a history-conditioned neural framework that integrates these sources of information through recurrent health representations and adaptive attribute conditioning. By distinguishing historical observation intervals from prospective prediction horizons, the framework moves beyond predicting the next observed health state to directly estimating coherent, horizon-specific transition probability distributions that can be aggregated into actuarial multi-state transition matrices.

The empirical results demonstrate the feasibility and competitiveness of this approach using longitudinal Health and Retirement Study data. Across one-, two-, and three-year forecast horizons, \lantern{} achieved competitive probabilistic accuracy, calibration, and aggregate transition-matrix estimation. At the primary two-year horizon, it outperformed multinomial logistic regression and the actuarial CTMC in multiclass probabilistic accuracy, while performing comparably to the strongest machine-learning benchmarks. These findings show that incorporating accumulated longitudinal information within an explicitly horizon-conditioned framework can provide reliable transition-probability estimates, although the independent contributions of individual architectural components remain to be established.

The principal contribution of our model therefore lies not solely in predictive improvements, but in providing a unified framework that connects history-dependent neural estimation with the horizon-specific probability structure required in actuarial multi-state modeling. Importantly, the study establishes the accuracy of transition estimates at observed horizons rather than validating their repeated use in long-term cohort projections. Future research should extend the framework to other longitudinal populations and insurance data environments, investigate history-updating projection methods that account for evolving individual characteristics, and examine how uncertainty in estimated transition probabilities propagates into projected disability occupancy and insurance liabilities. These developments would help establish the framework's practical value for long-term care risk assessment and actuarial valuation.

\section*{CRediT authorship contribution statement}
{\bf Bright Kwaku Manu}: Conceptualization, Methodology, Software, Formal analysis, Investigation, Data curation, Visualization, Writing -- original draft, Writing -- review \& editing; {\bf Beckett Sterner:} Conceptualization, Methodology, Supervision, Project administration, Writing -- review \& editing; {\bf Petar Jevti\'c:} Conceptualization, Methodology, Supervision, Project administration, Writing -- review \& editing.

\section*{Data and Code Availability}
The data used in this study are publicly available from the RAND HRS data files (RAND Center for the Study of Aging) at \url{https://hrsdata.isr.umich.edu/data-products/rand}. All code used to produce the results and figures in this work is available at \url{https://github.com/BrightManu-lang/lantern-health-prediction}.

\section*{Declaration of competing interest}
The authors declare that they have no known competing financial interests or personal relationships that could have appeared to influence the work reported in this paper.

\section*{Acknowledgments}
This work was supported by National Institutes of Health (NIH) grant 5R01GM131405-02, with Petar Jevti\'{c} and Beckett Sterner serving as primary investigators.

\medskip

\clearpage
\bibliographystyle{elsarticle-harv}
\bibliography{ref}

\clearpage

\section*{Supplementary Empirical Analyses}
\label{app:supplementary_results}

\renewcommand{\thetable}{S\arabic{table}}
\setcounter{table}{0}
\renewcommand{\thefigure}{S\arabic{figure}}
\setcounter{figure}{0}

The supplementary analyses provide additional evidence on horizon-specific calibration, uncertainty in predictive comparisons, learned attribute-attention patterns, and an illustrative actuarial application of estimated transition probabilities. A detailed dictionary of the RAND HRS variables is also provided.

\subsection*{Calibration Across Forecast Horizons}
\label{app:calibration}

\begin{figure}[h]
\centering
\includegraphics[width=\linewidth]{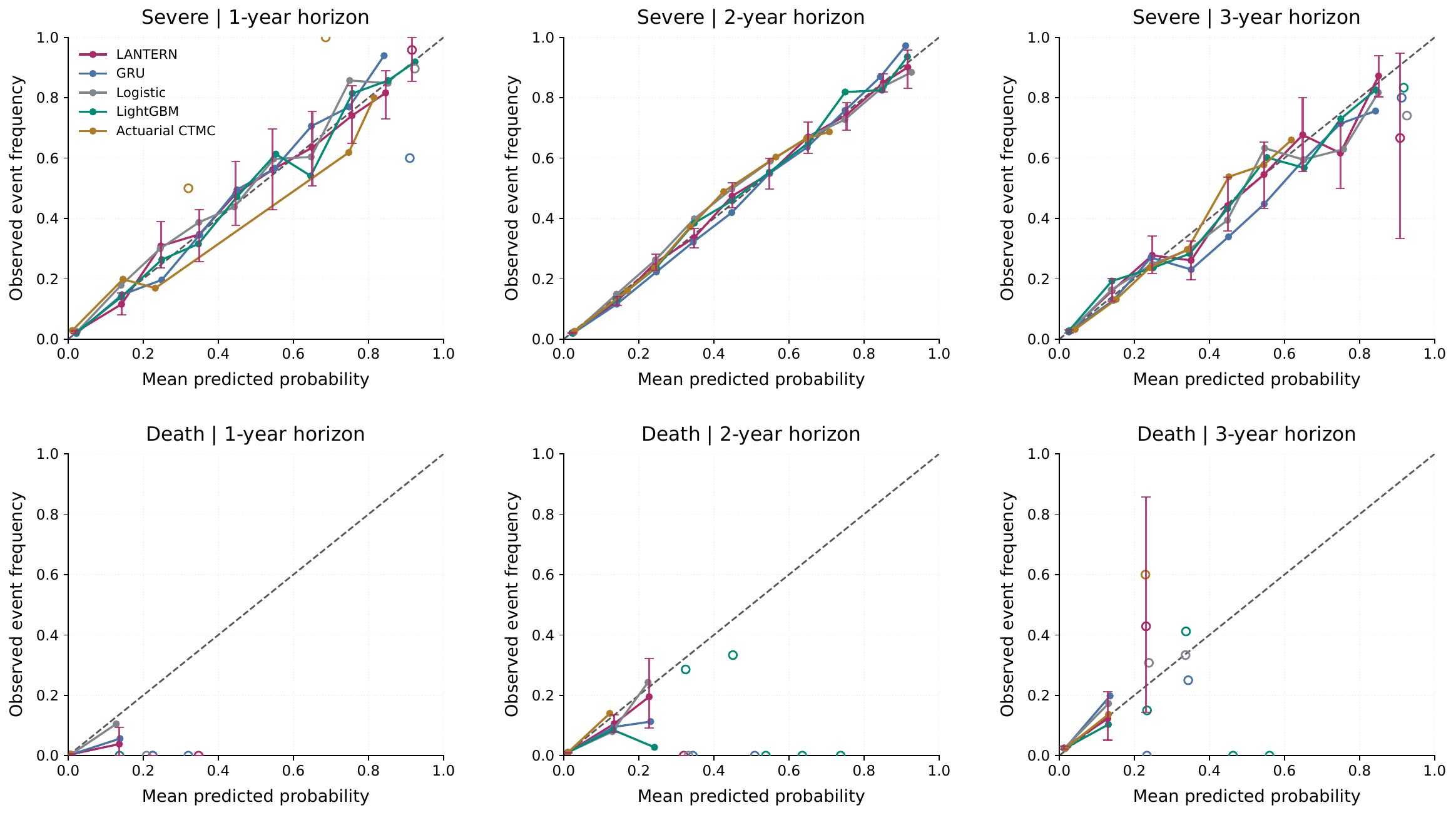}
\caption{Calibration of predicted severe-disability and mortality probabilities across one-, two-, and three-year forecast horizons. Points represent observed event frequencies against mean predicted probabilities within risk intervals for LANTERN and the benchmark models. The diagonal indicates perfect calibration. Vertical error bars represent 95\% individual-cluster bootstrap confidence intervals for LANTERN where available.}
\label{fig:calibration_all_horizons}
\end{figure}

Figure~\ref{fig:calibration_all_horizons} presents calibration diagnostics for severe disability and mortality across the one-, two-, and three-year forecast horizons. Severe-disability probabilities generally exhibit agreement between predicted and observed event frequencies, with greater variability in some probability intervals at the one- and three-year horizons. Mortality calibration is less informative at higher predicted risks because relatively few observations fall within these intervals. In particular, the one-year mortality evaluation contains only 21 observed deaths. These results complement the primary two-year calibration analysis and highlight the limited empirical support for assessing higher mortality risks.

\subsection*{Uncertainty in Predictive Performance Comparisons}
\label{app:predictive_bootstrap}

\begin{figure}[h]
\centering
\includegraphics[width=\linewidth]{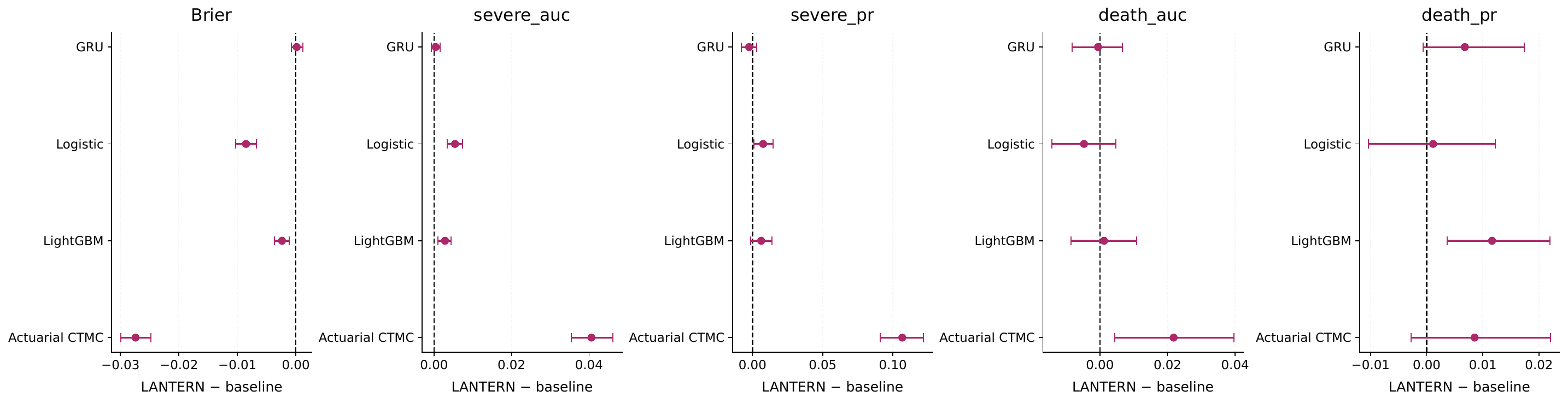}
\caption{Paired predictive-performance differences between LANTERN and the benchmark models at the two-year forecast horizon. Points represent LANTERN minus benchmark differences, with horizontal intervals indicating 95\% individual-cluster bootstrap confidence intervals. Negative differences favor LANTERN for the Brier score, whereas positive differences favor LANTERN for AUROC and average precision. The dashed vertical line indicates zero difference.}
\label{fig:predictive_bootstrap_tau2}
\end{figure}

Figure~\ref{fig:predictive_bootstrap_tau2} presents paired individual-cluster bootstrap comparisons of LANTERN and the benchmark models at the primary two-year horizon. The analysis considers differences in multiclass Brier score, severe-disability AUROC and average precision, and mortality AUROC and average precision. LANTERN exhibits a lower Brier score than LightGBM and higher mortality average precision, with confidence intervals excluding zero. However, several other comparisons, particularly those involving mortality AUROC, have intervals spanning zero. The results therefore indicate that the evidence for improved predictive performance depends on the endpoint, metric, and benchmark model.

\subsection*{Attribute-Attention Diagnostics}
\label{app:attention}

\begin{figure}[h]
\centering
\includegraphics[width=0.8\linewidth]{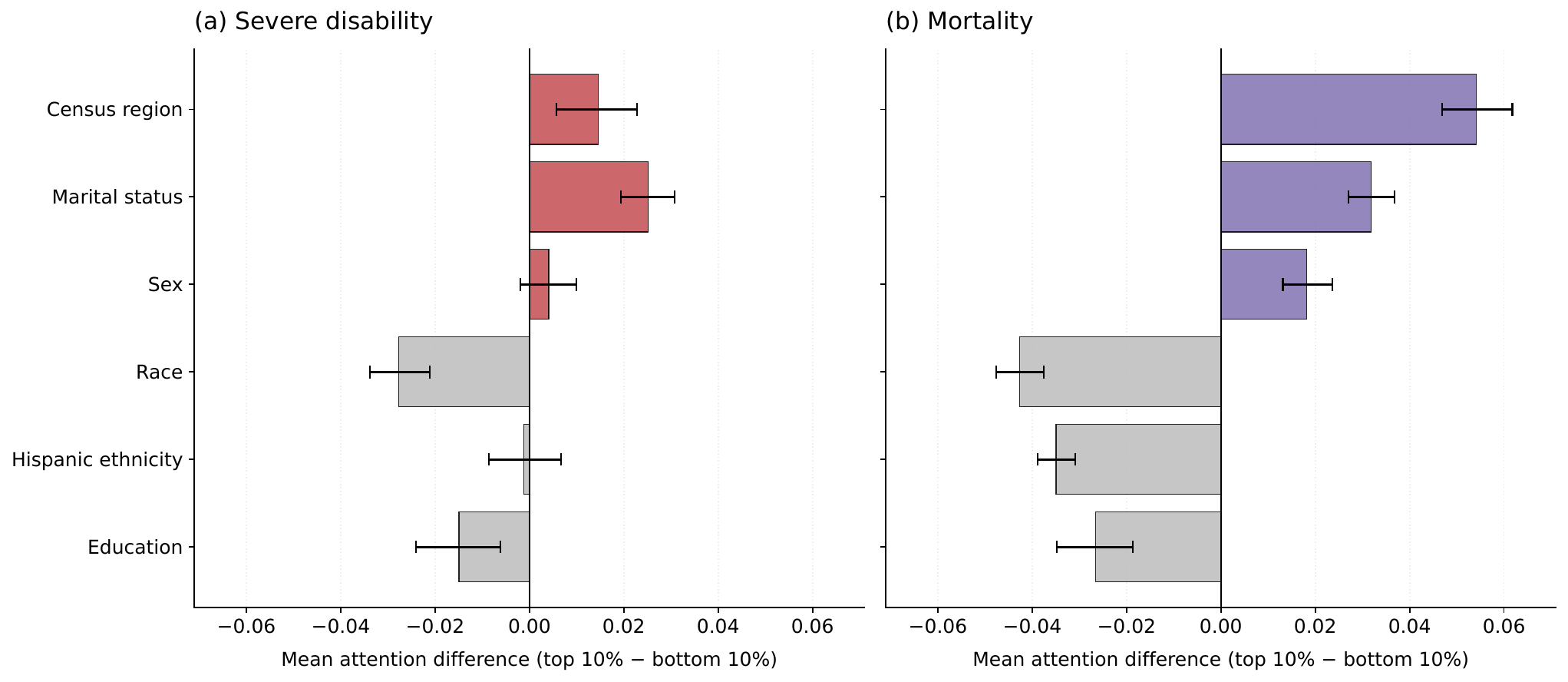}
\caption{Differences in learned attribute-attention weights between the highest and lowest predicted-risk deciles at the two-year horizon. Panel (a) presents severe-disability risk and panel (b) presents mortality risk. Bars represent differences in mean attention weights (top 10\% minus bottom 10\%), with horizontal whiskers indicating 95\% individual-cluster bootstrap confidence intervals. Positive differences indicate greater relative attention in high-risk observations.}
\label{fig:attention_shift_tau2}
\end{figure}

\subsection*{Illustrative Cohort Projection}
\label{app:cohort_projection}

To illustrate a possible actuarial application of the estimated transition probabilities, we consider a hypothetical cohort of 10,000 individuals initially classified as Healthy at age 60. Age-specific two-year transition matrices from LANTERN and the benchmark models are applied successively through age 90 to obtain illustrative cohort occupancy probabilities.

\begin{figure}[h]
\centering
\includegraphics[width=0.8\linewidth]{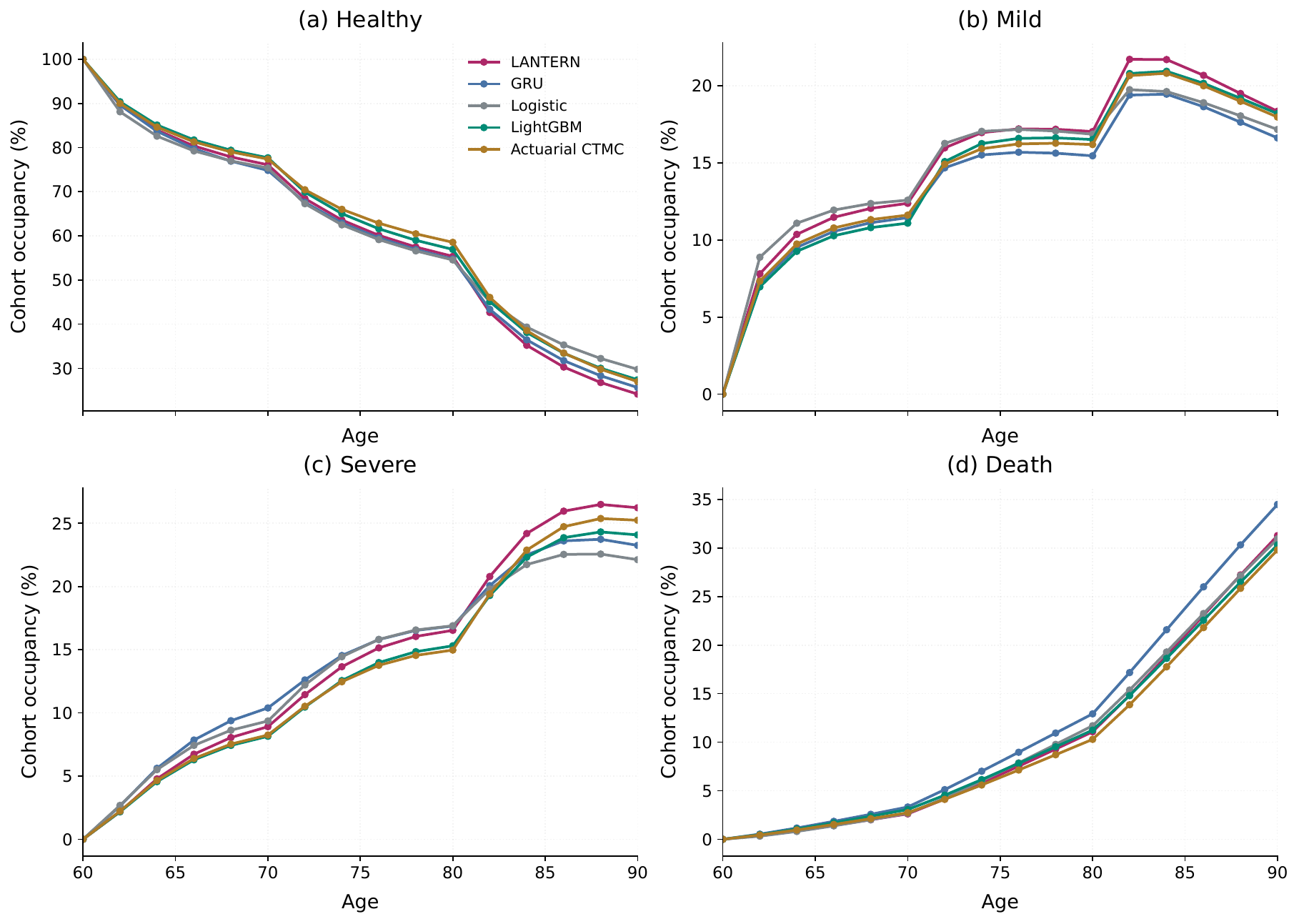}
\caption{Illustrative cohort occupancy trajectories obtained by repeated application of estimated age-specific two-year transition matrices. A hypothetical cohort of 10,000 individuals starts in the Healthy state at age 60 and is projected through age 90. Panels show the proportions occupying (a) Healthy, (b) Mild disability, (c) Severe disability, and (d) Death. Results are based on aggregated transition matrices and do not constitute validated long-term forecasts.}
\label{fig:cohort_projection}
\end{figure}

Figure~\ref{fig:cohort_projection} shows the resulting proportions occupying the Healthy, Mild disability, Severe disability, and Death states. The models produce broadly similar trajectories, although differences emerge in projected disability and mortality occupancy at older ages. Changes in the trajectories near age-group boundaries reflect the use of piecewise age-specific transition matrices.

These projections are illustrative rather than validated long-term forecasts. In particular, the repeated application of aggregated transition matrices assumes that the estimated transition probabilities remain applicable as the cohort evolves, without explicitly updating the distribution of individual health histories or other covariates. This assumption is restrictive for LANTERN because its transition probabilities depend on accumulated individual histories. Accordingly, the exercise demonstrates how differences in estimated transition matrices may translate into hypothetical cohort occupancy patterns, but does not establish the accuracy of long-term disability projections or actuarial valuations.

\newpage

\subsection*{Distribution of Retained Visits per Respondent}
\begin{table}[htbp]
\centering
\caption{Distribution of retained visits per respondent in the analytic sample.}
\label{tab:visit_distribution}

\begin{tabular}{lrr}
\hline
\textbf{Number of visits} &
\textbf{Respondents} &
\textbf{Percent (\%)} \\
\hline
2  & 3,622 & 9.60 \\
3  & 3,823 & 10.14 \\
4  & 5,041 & 13.37 \\
5  & 2,643 & 7.01 \\
6  & 2,997 & 7.95 \\
7  & 5,012 & 13.29 \\
8  & 2,025 & 5.37 \\
9  & 1,984 & 5.26 \\
10 & 3,014 & 7.99 \\
11 & 1,592 & 4.22 \\
12 & 1,516 & 4.02 \\
13 & 2,157 & 5.72 \\
14 & 2,232 & 5.92 \\
15 & 55    & 0.15 \\
\hline
\textbf{Total} & \textbf{37,713} & \textbf{100.00} \\
\hline
\end{tabular}

\vspace{0.3em}

\begin{minipage}{0.8\textwidth}
\footnotesize \scriptsize
\textit{Note:} Visit counts are calculated from retained health-state observations before the final transition eligibility filter and are restricted to respondents contributing at least one valid transition. The mean number of visits is 7.03 (SD = 3.65), with a median of 7 and an interquartile range of 4--10 visits.
\end{minipage}

\end{table}


\subsection*{Age-Specific Transition Counts and Follow-Up}

\begin{table}[htbp]
\centering
\caption{Age-specific transition counts and observed inter-visit
follow-up in the analytic sample.}
\label{tab:age_exposure}

\small
\begin{tabular}{lrrrr}
\hline
\textbf{Origin age} &
\textbf{Respondents} &
\textbf{Transitions} &
\textbf{Follow-up years} &
\textbf{Deaths} \\
\hline
30--39  & 396    & 821    & 1,764   & 0 \\
40--49  & 2,909  & 6,357  & 13,438  & 8 \\
50--59  & 21,000 & 60,106 & 125,298 & 179 \\
60--69  & 21,178 & 71,156 & 148,593 & 490 \\
70--79  & 17,681 & 57,894 & 121,724 & 896 \\
80--89  & 9,891  & 27,167 & 56,849  & 1,030 \\
90--100 & 1,969  & 3,869  & 7,887   & 337 \\
\hline
\textbf{Total} & \textbf{37,713$^{*}$} &
\textbf{227,370} & \textbf{475,553} &
\textbf{2,940} \\
\hline
\end{tabular}

\vspace{0.3em}

\begin{minipage}{0.8\textwidth}
\footnotesize \scriptsize
\textit{Note:} Follow-up years represent summed inter-visit intervals, not exact actuarial person-years at risk. Respondent counts across age groups are non-additive because individuals may contribute to multiple groups.
\end{minipage}

\end{table}


\subsection*{RAND HRS Variable Definitions}
\label{app:data_dictionary}

\begin{table}[p]
\centering
\small

\rotatebox{-90}{%
\begin{minipage}{\textheight}
\centering

\caption{RAND HRS variables and construction of analysis covariates}
\label{tab:rand_variables}

\begin{tabularx}{\textheight}{@{}l l X X@{}}
\hline
\textbf{Construct} & \textbf{RAND variable} & \textbf{Measurement / coding} & \textbf{Role in analysis} \\
\hline
Age at interview & AGEY\_E / AGEY\_B & Age in completed years & Time index and elapsed-time calculation \\
Activities of daily living & ADL6A & Limitation count, 0--6 & Longitudinal health feature and state construction \\
Instrumental activities of daily living & IADL5A & Limitation count, 0--5 & Longitudinal health feature and state construction \\
Self-rated health & SHLT & 1 = Excellent; 2 = Very good; 3 = Good; 4 = Fair; 5 = Poor & Longitudinal health feature \\
Cognition & COG27 & Score, 0--27 & Longitudinal health feature \\
Depressive symptoms & CESD & Eight-item score, 0--8 & Longitudinal health feature \\
Body mass index & BMI & Weight (kg)/height (m)$^2$ & Longitudinal health feature \\
Diabetes & DIABE & 0 = No; 1 = Yes & Chronic-condition feature \\
Cancer & CANCRE & 0 = No; 1 = Yes & Chronic-condition feature \\
Lung disease & LUNGE & 0 = No; 1 = Yes & Chronic-condition feature \\
Heart disease & HEARTE & 0 = No; 1 = Yes & Chronic-condition feature \\
Stroke & STROKE & 0 = No; 1 = Yes & Chronic-condition feature \\
Psychiatric condition & PSYCHE & 0 = No; 1 = Yes & Chronic-condition feature \\
Disease burden & Derived & Sum of six condition indicators, 0--6 & Longitudinal health feature \\
Marital status & MSTAT & 1 = Married; 2 = Spouse absent; 3 = Partnered; 4 = Separated; 5 = Divorced; 6 = Separated/divorced; 7 = Widowed; 8 = Never married & Wave-varying contextual attribute \\
Census region & CENREG & 1 = Northeast; 2 = Midwest; 3 = South; 4 = West; 5 = Other & Wave-varying contextual attribute \\
Sex & RAGENDER & 1 = Male; 2 = Female & Attribute-conditioning input \\
Race & RARACEM & 1 = White/Caucasian; 2 = Black/African American; 3 = Other & Attribute-conditioning input \\
Hispanic ethnicity & RAHISPAN & 0 = Not Hispanic; 1 = Hispanic & Attribute-conditioning input \\
Education & RAEDUC & 1 = Less than high school; 2 = GED; 3 = High-school graduate; 4 = Some college; 5 = College and above & Attribute-conditioning input \\
Age at death & RADAGE\_Y & Age at death in completed years & Death-state construction \\
\hline
\end{tabularx}

\end{minipage}
}
\end{table}

\end{document}